\documentclass[twoside]{article}

\usepackage{url}
\usepackage{booktabs}
\usepackage{amsfonts}
\usepackage{nicefrac}
\usepackage{microtype}
\usepackage{xcolor}
\usepackage{bm}

\usepackage{hyperref}
\hypersetup{
    colorlinks=true,
    linkcolor=blue,
    citecolor=blue,
    filecolor=magenta,      
    urlcolor=cyan,
    pdftitle={MixMIL},
    pdfpagemode=FullScreen,
}
\usepackage{amsmath}
\usepackage{graphicx}
\usepackage{caption}
\usepackage{ulem}
\usepackage{bbold}
\usepackage{enumitem}
\usepackage{float}
\usepackage{subcaption}
\usepackage{xr}

\usepackage{tocloft}
\usepackage[accepted]{aistats2024}

\usepackage[round]{natbib}

\long\def\cut#1{}

\newcommand{\be}{\begin{equation}}
\newcommand{\ee}{\end{equation}}
\newcommand{\bea}{\begin{eqnarray}}
\newcommand{\eea}{\end{eqnarray}}
\newcommand{\beas}{\begin{eqnarray*}}
\newcommand{\eeas}{\end{eqnarray*}}

\newcommand{\given}{\,|\,}

\newcommand{\B}[1]{\bm{#1}}

\begin{document}
\runningauthor{Engelmann, Palma, Tomczak, Theis, Casale}

\twocolumn[

\aistatstitle{Mixed Models with Multiple Instance Learning}

\aistatsauthor{Jan P. Engelmann\textsuperscript{*}
 \And Alessandro Palma\textsuperscript{*}}

\aistatsaddress{ Helmholtz Munich 
\And  Helmholtz Munich } 

\aistatsauthor{Jakub M. Tomczak \And Fabian J. Theis\textsuperscript{\dag} \And Francesco Paolo Casale\textsuperscript{\dag}}
\aistatsaddress{Eindhoven University of Technology \And
Helmholtz Munich \And Helmholtz Munich }
]

\begin{abstract}
Predicting patient features from single-cell data can help identify cellular states implicated in health and disease. Linear models and average cell type expressions are typically favored for this task for their efficiency and robustness, but they overlook the rich cell heterogeneity inherent in single-cell data. To address this gap, we introduce MixMIL, a framework integrating Generalized Linear Mixed Models (GLMM) and Multiple Instance Learning (MIL), upholding the advantages of linear models while modeling cell state heterogeneity. By leveraging predefined cell embeddings, MixMIL enhances computational efficiency and aligns with recent advancements in single-cell representation learning. Our empirical results reveal that MixMIL outperforms existing MIL models in single-cell datasets, uncovering new associations and elucidating biological mechanisms across different domains.
\end{abstract}

\section{INTRODUCTION}
Single-cell omics data have been instrumental in unveiling cellular heterogeneity, proving invaluable in studying human health and disease~\citep{perez2022single,ahern2022blood,vandereyken2023methods,lim2023transitioning}. Within these vast datasets, determining which cells are impacted by specific interventions or genetic variations is of utmost importance. However, drawing such associations at the single-cell level is statistically challenging, given the sparse nature of the data and the structured noise introduced by cellular dependencies within a sample~\citep{you2023modeling,cuomo2023single}. Traditional pooling procedures, such as the pseudo-bulk approach in single-cell RNA sequencing, sidestep these challenges but at the expense of overlooking key cell states~\citep{perez2022single,ahern2022blood,YazarOneK1K,janssens2021fully,lafarge2019capturing,ljosa2013comparison}.

Multiple instance learning provides a principled framework for modeling cellular heterogeneity through attention and has proven effective in several domains~\citep{Ilse2018attentionmil,additive_mil,ds_mil,bayes_mil, Shao2021transMIL}.
Yet, its application to single-cell datasets has been limited. This might be attributed to the characteristic low signal-to-noise ratio of these datasets, a setting where simple models based on mean pooling and linear assumptions tend to perform adequately \citep{Crowell2020}.

To bridge this gap, we introduce Mixed Models with Multiple Instance Learning (MixMIL), a new framework integrating the robustness of the Generalized Linear Mixed Model (GLMM) with the MIL ability to model heterogeneity.
Designed for robustness and efficiency in single-cell analyses, MixMIL leverages cell embeddings from pre-trained unsupervised models~\citep{Theodoris2023,Cui2023,Doron2023} and integrates a simple attention-based MIL module into GLMMs, synthesizing the strengths of both frameworks.
In extensive simulations and evaluations, we benchmarked MixMIL against the GLMM and state-of-the-art MIL architectures. Across a spectrum of applications, spanning single-cell genomics to microscopy and reaching into histopathology, MixMIL consistently outperformed other MIL implementations, underscoring that reduced complexity through principled model design can notably enhance results in these contexts.

\renewcommand*{\thefootnote}{\fnsymbol{footnote}}
\footnotetext{\raggedright{$^{*}$Equal contribution. \textsuperscript{\dag}Correspondence to: \{francescopaolo.casale, fabian.theis\}@helmholtz-munich.de}}
\renewcommand*{\thefootnote}{\arabic{footnote}}
\setcounter{footnote}{0}
%
\section{RELATED WORK}\label{sec:relwork}
\paragraph{Multiple Instance Learning}
MixMIL models cellular heterogeneity within the GLMM framework using MIL.
MIL organizes data into collections of instances, called \textit{bags}, and operates on the premise that the label of the collection is known while the labels of individual instances remain unknown. In attention-based MIL (ABMIL), \cite{Ilse2018attentionmil} trained a bag-level classification model where a neural network learns the influence of single instances on the bag label prediction in the form of attention weights. Later efforts built upon the notion of attention-based deep MIL to improve the interpretability of instance contribution~\citep{additive_mil} and alleviate performance degradation due to class imbalance~\citep{ds_mil}. More related to our setting, \citet{bayes_mil} used variational inference to estimate the posterior distribution of instance-level weights to enhance model interpretability and uncertainty estimation. In the medical field, MIL has been widely employed to study the heterogeneous effect of disease in large histopathology images in the context of cancer prediction \citep{Ilse2018attentionmil, Sudharshan2019histopath_class, Zhao2020histopath_class,wagnerMIL}, segmentation~\citep{Lerousseau2020histopath_seg} and somatic variant detection~\citep{Cui2020somatic}.
In contrast to earlier approaches, MixMIL uniquely integrates attention-based MIL within GLMMs.
\vspace{-\parskip}
\paragraph{Generalized Linear Mixed Models}
MixMIL introduces multiple instance learning within the GLMM framework.
GLMMs extend linear models with random effects to enable robust regression and hierarchical modeling, and they are equipped to handle a variety of outcome distributions~\citep{breslow1993approximate,bates2014fitting}.
Specialized GLMMs have become indispensable in genomics, especially in association analysis~\citep{lippert2011fast,bates2014fitting,loh2018mixed} and interaction testing~\citep{casale2017joint,moore2019linear,dahl2020robust}.
In recent advancements, GLMM-based interaction tests have been employed to model cell state heterogeneity in single-cell datasets~\citep{neavin2021single,cuomo2022cellregmap,gewirtz2022expression,nathan2022single}.
These tests primarily focus on associating singular patient and genomic features while modeling effect heterogeneity—for instance, exploring how the effect of a genetic variant might regulate the expression of an individual gene based on cell state~\citep{cuomo2022cellregmap}.
In contrast, MixMIL seeks to characterize single patient features using the cell state representations from a group of cells, uniquely incorporating a MIL module in GLMMs for this purpose.
\paragraph{Predefined Embeddings and Single-Cell Atlases}
MixMIL employs shallow machine learning functions on predefined embeddings for robustness and efficiency.
This synergy was first spotlighted in representation learning for computer vision with frameworks such as SimCLR~\citep{Chen2020SimCLR} and later extended by others~\citep{He2020Moco,Caron2021DINO}.
Rapidly, this influence radiated across biological disciplines, advancing representational learning in computational pathology~\citep{BenTaieb2020KimiaNet,wang2023retccl}, single-cell genomics~\citep{Lopez2018scVI,Theodoris2023, Cui2023}, and microscopy~\citep{MarinZapata2020dino, Siegismund2022archetypes}.
While the use of predefined embeddings in MIL has been considered elsewhere \citep{ds_mil, Shao2021transMIL}, the synergy with MixMIL is especially timely within the single-cell omics sphere.
As the domain leans towards foundational models for comprehensive single-cell atlases~\citep{schiller2019human, travaglini2020molecular, deprez2020single, wagner2019single, wilk2020single, sikkema2022integrated}, MixMIL stands primed, ready to leverage the wealth of emerging high-quality embeddings, equipping researchers for robust, integrated analyses.

\section{METHODOLOGY}

\subsection{Problem Statement}
Multiple instance learning is a variation of supervised learning where the training set consists of labeled bags, each containing several instances.
Formally, a bag associated with a single label $y$ consists of $I$ unordered and independent instances $\{\B{x}_1, \dots, \B{x}_I\}$, where $\B{x}_i\in\mathbb{R}^Q$.
We here collectively denote these instances with $\B{X}=\left[\B{x}_1,\dots,\B{x}_I\right]^T\in \mathbb{R}^{I \times Q}$.
Notably, the number of instances $I$ can vary across different bags.
The primary goal of MIL is to predict the label $y$ from the bag of instances $\B{X}$, using a function that is invariant to permutations among instances. This problem could be solved by a model that defines instance embeddings through a function $f$, $f(\B{X}) = \{f(\B{x}_{1}), \ldots, f(\B{x}_{I})\}$, and then aggregates them into a single bag embedding $\B{z}$, which is eventually fed to a predictor.
In the following, we focus on two approaches that will serve as an inspiration for our method.
First, we present the attention-based deep MIL framework~\citep{Ilse2018attentionmil} that utilizes deep neural networks and the attention mechanism for aggregation.
Second, we outline a GLMM~\citep{breslow1993approximate}, an extension of linear models widely used in genomic analyses.
\vspace{-\parskip}
\subsection{Attention-Based MIL}
\cite{Ilse2018attentionmil} introduced an innovative pooling function for MIL, implementing the concept of attention to aggregate instance-level features into bag-level ones.
Specifically, they first introduce a neural network function $f$ to derive low-dimensional instance embeddings from instance features $\B{x}_i$ and then use a weighted average pooling function equivalent to the \textit{attention mechanism} to aggregate $\{f(\B{x}_1), \dots, f(\B{x}_I)\}$ into bag embeddings $\B{z}$. Finally, they consider a classifier to predict the bag label $y$ from $\B{z}$. The weighting function was defined as follows:
\begin{equation}
\B{z}=f(\B{X})^T\B{w},\;\;\text{with $w_i>0$ $\forall{i}$ and $\Sigma_iw_i=1$},
\label{eq:ilse}
\end{equation}
where $\B{w}\in\mathbb{R}^{I}$ denotes the vector of importance weights across the $I$ instances. Modeling the importance weights as a two-layer neural network function with softmax activation function on the last layer, $\omega(f(\B{X}))$, they can ensure that the constraints on the weights are always satisfied while the entire architecture can be trained end-to-end---i.e., the function $f$, the importance weight function $\omega$ and the bag-level classifier can be jointly optimized.

Following the foundational work by \citet{Ilse2018attentionmil}, more advanced implementations of attention-based MIL have emerged.
For example, DSMIL computes attention weights for each observation in a bag based on their similarity with the instance which has the highest classification score for a certain class ~\citep{ds_mil}. Attention weights are then used to aggregate features. In multiclass classification, each class has its own critical instances, which allows DSMIL to use different importance weighting for different classes. 
Moreover, \citet{bayes_mil} suggested using Bayesian neural networks for attention-based MIL. Such an approach optimizes a posterior on the parameters of the attention function via variational inference and yields calibrated uncertainties for better weight interpretability.

\subsection{Generalized Linear Mixed Model for MIL}
We can employ a GLMM in the context of MIL to model the relationship between the bag label $y$ and fixed bag embeddings $\B{z}(\B{X})$ derived from bag $\B{X}$ while accounting for bag covariates $\B{c}$.
Specifically, given a link function $g$, the expected value of the bag label, $\mu = \mathbb{E}[y|\B{X}]$, is linked to a linear predictor of bag embeddings $\B{z}(\B{X})$ and covariates $\B{c}$ through
%
\begin{equation}
g(\mu) = \B{c}^T\B{\alpha} + \B{z}(\B{X})^T\B{\beta},
\label{eq:glmm}
\end{equation}
where $\B{\alpha}\in\mathbb{R}^{K}$ and $\B{\beta} \in \mathbb{R}^{Q}$ denote the effects of the $K$ covariates and $Q$-dimensional bag embeddings, respectively.
In the case of high-dimensional bag embeddings, $\B{\beta}$ is modeled as a random effect to improve robustness.
We note that for average pooled bag embeddings, which are common in single-cell omics analysis~\citep{perez2022single,ahern2022blood,YazarOneK1K,janssens2021fully,lafarge2019capturing,ljosa2013comparison}, we have $\B{z}(\B{X}) = \frac{1}{I}\sum_{i=1}^{I} f(\B{x}_i)$.

\subsection{Our Approach: The MixMIL Model}

\begin{figure*}[h]
\begin{centering}
\includegraphics[width=0.95\textwidth]{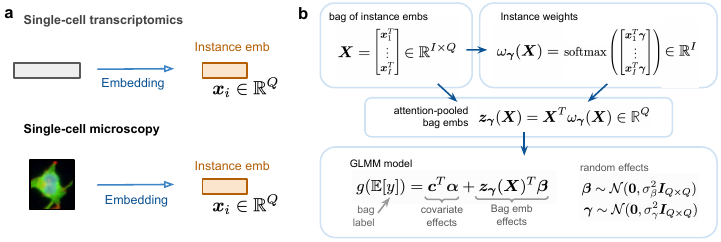}
\vskip -2mm
\captionsetup{font=small}
\captionsetup{skip=20pt}
\caption{
(\textbf{a}) MixMIL uses predefined instance embeddings from domain-specific unsupervised models for robustness and efficiency.
(\textbf{b}) Generalized multi-instance mixed model framework defining MixMIL.
}
\label{fig:method}
\end{centering}
\end{figure*}

We propose a novel integration of the attention-based MIL and GLMM frameworks, as illustrated in Figure~\ref{fig:method}. Our approach encompasses two main steps:
(i) Utilize predefined instance embeddings and model the importance weights as a shallow function of them;
(ii) Replace the static pooling function in standard GLMMs~\eqref{eq:glmm} with a dynamic, trainable attention-based pooling function that leverages the aforementioned importance weights.

\subsubsection{The Model}
%

\paragraph{Predefined Embeddings and Shallow Attention Weight Function}
Leveraging insights from recent advances in representation learning~(see Section~\ref{sec:relwork}), we employ predefined instance embeddings from domain-specific unsupervised models as instance features, bypassing the need for end-to-end optimization of a feature extractor $f$. These embeddings are ubiquitously available across various data modalities (see Section~\ref{sec:relwork}).
Additionally, we model instance importance weights using a single linear layer with a softmax activation function across instances.
With these assumptions, the bag embeddings can be written as follows:

{\small
\begin{equation}\label{eq: bag_embeddings}
\B{z}_{\B{\gamma}}(\B{X})=\B{X}^T\omega_{\B{\gamma}}(\B{X})\in\mathbb{R}^{Q}, 
\end{equation}
\begin{equation}\label{eq: bag_embeddings_softmax}
\text{with  $\omega_{\B{\gamma}}(\B{X})=
\text{softmax}\left(\B{X}\B{\gamma}\right)=
\text{softmax}\left(
{\tiny
\begin{bmatrix}
\B{x}_1^T\B{\gamma} \\
\vdots \\
\B{x}_I^T\B{\gamma}
\end{bmatrix}
}
\right)\in\mathbb{R}^{I}$}, \nonumber
\end{equation}
\small}
where $\B{X}\in\mathbb{R}^{I\times{Q}}$ denotes the predefined embeddings across all instances, and we made explicit that both the bag embeddings $\B{z}$ and the weight function $\omega$ depend on the parameters $\B{\gamma}$. This way of aggregating bag embeddings is an \textit{attention mechanism} as defined by \cite{Ilse2018attentionmil} in Eq.~\eqref{eq:ilse}. 

\paragraph{Modeling Dependencies} 
To model the relationship between the bag label $y$ and bag embeddings $\B{z}_{\B{\gamma}}(\B{X})$ defined in Eq.~\eqref{eq: bag_embeddings}, we consider the GLMM formulation for MIL in Eq.~\eqref{eq:glmm}:
%
\begin{equation}
g(\mu)=\B{c}^T\B{\alpha}+\B{z}_{\B{\gamma}}(\B{X})^T\B{\beta},
\label{eq:logits}
\end{equation}
where now the bag pooling function $\B{z}_{\B{\gamma}}(\B{X})$ (specified in Eq.~\eqref{eq: bag_embeddings}) is dynamic and end-to-end trainable.
To ensure robust regression for small sample sizes or higher-dimensional instance embeddings, we model both $\B{\beta}$ and $\B{\gamma}$ as random effects, i.e., we introduce the priors $\B{\beta}\sim\mathcal{N}(\B{0}, \sigma_{\beta}^2\B{I}_{Q\times{Q}})$ and $\B{\gamma}\sim\mathcal{N}(\B{0}, \sigma_{\gamma}^2\B{I}_{Q\times{Q}})$. Here, $\B{I}_{Q\times{Q}}$ denotes the $Q\times{Q}$ identity matrix, and $\sigma_{\beta}^2$ and $\sigma_{\gamma}^2$ are the variances associated with the parameters $\B{\beta}$ and $\B{\gamma}$.
Given these priors, namely:
%
\begin{align}
\text{p}\label{eq: beta_prior}(\B{\beta})=\mathcal{N}\left(\B{\beta}\given\B{0}, \sigma_{\beta}^2\B{I}_{Q \times Q}\right)\; \\ 
\text{p}\label{eq: gamma_prior}(\B{\gamma})=\mathcal{N}\left(\B{\gamma}\given\B{0}, \sigma_{\gamma}^2\B{I}_{Q \times Q}\right)\; ,
\end{align}
the marginal likelihood of the model is the following:
{\small
\begin{equation}\label{eq: likelihood}
\text{p}(y|\B{c},\B{\alpha},\B{X})= 
\int
\text{p}\left(y\left|\B{c}^T\B{\alpha}+\B{z}_{\B{\gamma}}(\B{X})^T\B{\beta}\right.\right)\text{p}(\B{\beta})\text{p}(\B{\gamma}) \text{d}\B{\beta}\text{d}\B{\gamma}
\end{equation}
\small}
This integral is generally intractable, but it can be approximated using various techniques such as Monte Carlo methods, Laplace approximation, or variational inference. In this work, we opt to use variational inference.\looseness=-1 

\paragraph{Instance Importance Heterogeneity}
The parameter $\sigma_{\gamma}^2$ primarily controls the heterogeneity of the importance weights across instances.
Indeed, when $\sigma_{\gamma}^2=0$, we have $\B{\gamma}=\B{0}$, and $\B{z}_{\B{0}}(\B{X})$ reduces to a simple average across all instances.
In this case, MixMIL simplifies to a standard GLMM with average pooled bag features. Conversely, larger values of $\sigma_{\gamma}^2$ correspond to a more significant disparity in importance weights across instances, with only a few instances contributing the most to bag label predictions.

\paragraph{Model Interpretability}
Given that both the predictor from bag embeddings and the pooling function are linear in the instance embeddings, the aggregate effect of bag embeddings on bag labels can be expressed as follows:
%
\begin{equation}\label{eq:interpretable}
\B{z}_{\B{\gamma}}(\B{X})^T\
\B{\beta}=
\omega_{\B{\gamma}}(\B{X})^T\B{X}\B{\beta}=
\omega_{\B{\gamma}}(\B{X})^Tt_{\B{\beta}}(\B{X}),
\end{equation}
where $t_{\B{\beta}}(\B{X})=\B{X}\B{\beta}\in\mathbb{R}^I$ can be viewed as the vector of instance-level phenotypic predictions.
This interpretable formula provides a way to decompose the overall bag prediction into a weighted sum of instance-level contributions, offering insights into the individual instance influences on the final prediction.

\subsubsection{Inference}
The aim in inference is to determine the posterior distribution $\text{p}(\B{\theta}\given\mathcal{D})$ of the random effect parameters $\B{\theta}=\{\B{\beta}, \B{\gamma}\}$ given the observed data $\mathcal{D}$. As exact inference is intractable for our model, we resort to variational inference, a strategy that approximates the true posterior $\text{p}(\B{\theta}\given\mathcal{D})$ by introducing a variational family $\text{q}_{\B{\phi}}(\B{\theta})$ parameterized by $\B{\phi}$, and optimizing $\B{\phi}$ to maximize the Evidence Lower Bound (ELBO):
%
{\small
\begin{align}
\text{ELBO}\bigl(\B{\phi}, \sigma^2_\beta, \sigma^2_\gamma\bigr) = \mathbb{E}_{\text{q}_{\B{\phi}}(\B{\theta})}\bigl[\log \text{p}(\mathcal{D} | \B{\theta})\bigr] - D_{\text{KL}}\bigl(\text{q}_{\B{\phi}} || \text{p}\bigr).
\end{align}}
Here $D_{\text{KL}}\bigl(q_{\B{\phi}} || \text{p}\bigr)$ denotes the Kullback-Leibler divergence between the variational approximation $\text{q}_{\B{\phi}}(\B{\theta})$ and the prior distribution of the parameters $\text{p}(\B{\theta})$.\looseness=-1 

We here consider the variational family of multivariate Gaussian distributions with full rank covariance:
%
\begin{equation}
\text{q}_{\B{\phi}}(\B{\theta})=\mathcal{N}\left(\B{\theta}\given\B{\mu}_{\B{\phi}},\B{\Sigma}_{\B{\phi}}\right),
\end{equation}
parameterized by $2Q$ mean parameters and $Q(2Q+1)$ covariance parameters.
In simulations, we also explore a mean field variational family, which assumes a fully factorized posterior across parameter dimensions---leading to $2Q$ mean parameters and $2Q$ variance parameters.

\paragraph{Optimization}
We jointly optimize the ELBO with respect to fixed effects $\B{\alpha}$, variational parameters $\B{\phi}$, and prior hyperparameters $\sigma_{\beta}^2$ and $\sigma_{\gamma}^2$ using mini-batch gradient descent. This optimization strategy, aligning with the empirical Bayes method~\citep{empirical_bayes}, adjusts the prior distributions in response to observed data. In order to backpropagate through the expectation term in the ELBO, we sample from the variational posterior and utilize the reparameterization trick~\citep{Ranganath2014bb_vi}. In our experiments, we found that using the Adam optimizer with a learning rate of $10^{-3}$, a training batch size of 64 bags, and 8 posterior samples to approximate the expectation, produced robust results across all settings. 

\subsubsection{Predictive Posterior}
After optimization, we employ the learned approximate posterior $\text{q}(\B{\beta}, \B{\gamma})$ to predict the label of a new bag from its instance embeddings $\B{X}^{\star}$:
%
\begin{equation}
\B{y}^\star=\mathbb{E}_{\text{q}(\B{\beta}, \B{\gamma})}\left[
\omega_{\B{\gamma}}(\B{X}^{\star})^Tt_{\B{\beta}}(\B{X}^{\star})\right],
\label{eq:prediction}
\end{equation}
where we used the formulation in Eq.~\eqref{eq:interpretable}.
Moreover, to retrieve important instances, we can leverage the expected value of the importance weights, namely $\mathbb{E}_{\text{q}(\B{\beta}, \B{\gamma})}\bigl[\omega_{\B{\gamma}}(\B{X}^{\star})\bigr].$

\subsubsection{Likelihood Choices in Experiments}
\label{sec:likelihood_choices}

\paragraph{Genotype Labels}
For genotype labels, where the label represents the minor allele count with possible values in the set \{0, 1, 2\}, we employ a Binomial likelihood with two trials~\citep{hao2016probabilistic}. The linear predictor operates in the logit space, and the resulting probability for the binomial distribution is derived from the sigmoid function applied to the logits.

\paragraph{Multiclass Classification}
For the multiclass classification problem in the microscopy dataset, we employ a categorical likelihood. Specifically, given $C$ classes, we use parameters $\bm{\alpha}\in\mathbb{R}^{K \times C}$, $\bm{\beta}\in\mathbb{R}^{Q \times C}$, and $\bm{\gamma}\in\mathbb{R}^{Q \times C}$ to specify class-specific covariate, feature, and attention effects.
The aggregated predictions as per Eq.~\eqref{eq:prediction} then produce a $C$-dimensional logit vector.
The class probabilities are then derived from the logit vector using the softmax link function.
With this approach, MixMIL can learn different attention mechanisms for different classes. We employ the same prior for feature and attention effects across all classes as outlined in Eq.~(\ref{eq: beta_prior}-\ref{eq: gamma_prior}).\looseness=-1 

\paragraph{Binary Classification}
For the histopathology classification task, we use a Bernoulli likelihood. The linear predictor operates in the logit space, and the probability is determined by the sigmoid function applied to the logits.

\subsubsection{Implementation and Complexity}

\paragraph{Implementation\footnote{\url{https://github.com/AIH-SGML/MixMIL}}}
To facilitate efficient training and inference on both GPU and CPU, we implemented MixMIL using PyTorch.
This choice also enabled us to leverage the numerous probability distributions already available in Pytorch for our generalized likelihood framework.
For efficient computation of bag-level operations across all bags (e.g., bag-level softmax), we utilized the PyTorch Scatter library. Importantly, our code supports the simultaneous analysis of multiple labels, which we make efficient by tensorizing computations across outcomes.

\paragraph{Model Size and Complexity}
MixMIL employs single linear layers for importance labels and predictions, resulting in $2Q+K$ likelihood parameters, where $Q$ represents the number of instance features and $K$ denotes the number of bag covariates. This is significantly fewer than MIL baselines like ABMIL, DSMIL, and BayesMIL (Table~\ref{table:genetics}). However, we note that MixMIL's variational posterior can notably exceed the likelihood's parameter count, especially with the multivariate Gaussian variational posterior having $2Q+Q(2Q+1)$ parameters.

\section{EXPERIMENTS}
We demonstrate the utility of MixMIL by applying it to the task of making predictions for unseen bags of instances. After benchmarking our model in extensive simulations, we selected three applications from diverse domains: (i) predict genetic labels from transcriptional cell embeddings, (ii) predict a compound's Mode of Action (MoA) from morphological cell embeddings, (iii) classify histology slides between cancer vs healthy from histological patch embeddings.

\subsection{Methods Considered}
\label{method_considered}

\paragraph{MIL Models}
We considered the established ABMIL~\citep{Ilse2018attentionmil} and its variation Gated ABMIL.
Additionally, we included recently published methods DSMIL~\citep{ds_mil} and Bayes-MIL~\citep{bayes_mil}.
For the MoA prediction tasks, we also considered Additive MIL~\citep{additive_mil}, given its reported improved performance in multiclass prediction tasks.
For full details on hyperparameter sweeping and selection across the different experiments, see Appendix~\ref{sec:experimental_setup}.

\vspace{-\parskip}
\paragraph{Traditional ML Models}
In simulation studies, we also benchmarked MixMIL against a GLMM with a corresponding likelihood and two widely used nonlinear models: Random Forest and XGBoost.
All models have predefined bag-level features as inputs. For this, we explored three different strategies: mean pooling, median pooling, and a combination of mean, squared mean, and cubed mean of instance features. Detailed implementation specifics and comparisons with such models on the simulation setting are provided in Appendix~\ref{single_cell_genomics_dataset} and \ref{sec:simulation_appendix}. We initially also tried to train these models directly on instance features by assigning bag labels to the corresponding instances. However, these models were too slow and underperformed (see Table~\ref{tab:instance-models}) and thus were excluded from further comparisons. 

\begin{figure*}[t!]
\begin{centering}
\includegraphics[width=0.9 \textwidth]{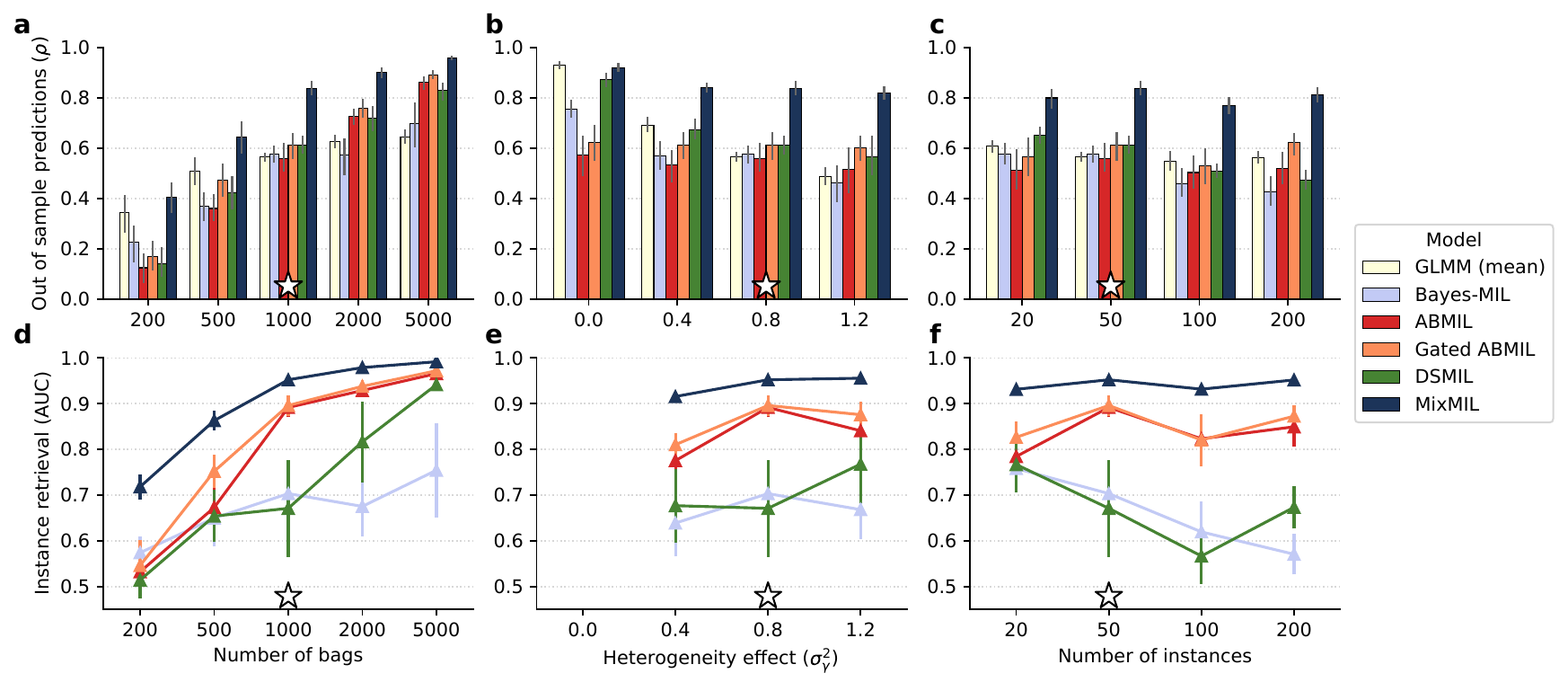}
\vskip -4mm

\captionsetup{font=small}
\captionsetup{skip=20pt}
\caption{
(a-c) Out-of-sample prediction accuracy (Spearman correlation, $\rho$) for MixMIL, GLMM, and baseline MILs (ABMIL, Gated ABMIL, DSMIL, and Bayes-MIL) varying the sample size (a), the amount of instance importance heterogeneity (b) and the number of instances (c).
(d-f) Instance retrieval ROC-AUC of MixMIL and baseline MILs for the top 10\% of instances in the same simulated scenarios. GLMM is not shown as it is not designed for instance retrieval.
Stars denote default values that were kept constant while varying other parameters. Error bars denote standard errors across 10 repeat experiments. Full results across all methods and scenarios can be found in Section~\ref{sec:simulation_appendix}.
}
\vskip -3mm
\label{fig:simulation}
\end{centering}
\end{figure*}
\subsection{Simulations}

\paragraph{Dataset Generation}
We designed our simulation to emulate the single-cell genomics application. First, we generated instance embeddings 
$\B{X}_{iq}\sim\mathcal{N}(0, 1)$
, importance weights $\omega_{\B{\gamma}}$ with
$\B{\gamma}_{q} \sim \mathcal{N}(0, \sigma_{\gamma}^2)$
, and bag embedding effects 
$\B{\beta}_{q}\sim\mathcal{N}(0, \sigma_{\beta}^2)$.
Next, we employed the model in Eq.~\eqref{eq:logits} to generate bag-level logits.
Finally, we generated genotypes using a binomial likelihood function with 2 trials taking as input the simulated logits (see also Section~\ref{sec:likelihood_choices}).
We systematically varied parameters such as the sample size, instance importance heterogeneity ($\sigma_{\gamma}^2$), the number of instances per bag, the number of instance features, and the variance explained by bag embedding effects. For each parameter configuration, we ran 10 repeat experiments.

\paragraph{Setup}
Out-of-sample prediction accuracy for all models was measured using Spearman's rank correlation between the actual bag logits and the predicted values in a simulated test set of 200 bags.
To evaluate the effectiveness of the MIL models to retrieve the most important instances using the weight posterior, we used ROC-AUC for the top 10\% of the simulated instances.
Standard errors on all metrics were computed across the 10 repeat experiments.

\begin{figure*}[!ht]
\begin{centering}
\includegraphics[width=0.8 \textwidth]{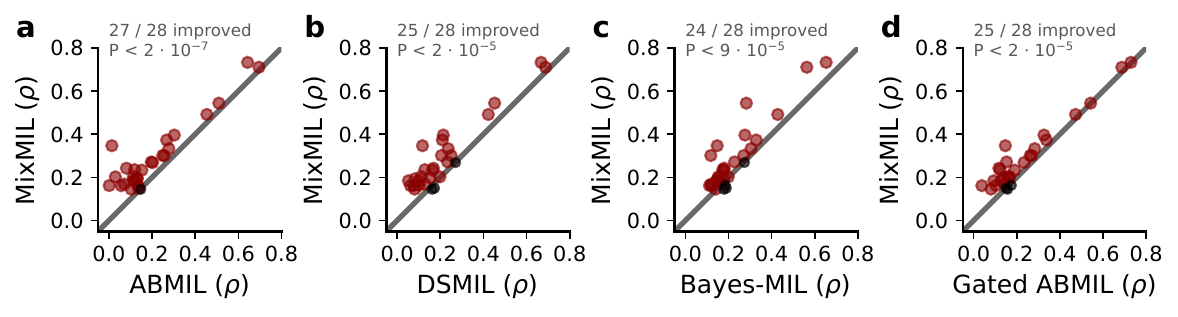}
\vskip -4mm
\captionsetup{font=small}
\captionsetup{skip=20pt}
    \caption{Scatter plots comparing the prediction performance (Spearman correlation, $\rho$) of MixMIL (y-axis) against baseline MILs (x-axis) for 28 genetic labels: MixMIL vs ABMIL (a), MixMIL vs DSMIL (b), MixMIL vs Bayes-MIL (c), MixMIL vs Gated~ABMIL (d). Genetic labels for which MixMIL yielded improved prediction accuracy are highlighted in red. The count of these genes and the P-values from a binomial test (assuming a null of 50/50 performance chance over 28 trials) are reported for each comparison.}
    \label{fig:genetics}
\end{centering}
\end{figure*}

\paragraph{Results}
When evaluating models across varying sample sizes, we noticed variable relative performance of the compared models: while the GLMM was superior to baseline MILs at lower sample sizes, baseline MILs gradually improved with more samples, all surpassing the GLMM at around 2,000 bags (Figure~\ref{fig:simulation}a). In contrast, MixMIL outperformed all baselines throughout (Figure~\ref{fig:simulation}a), emphasizing its reliability in settings where traditional MIL models might be prone to overfitting.
As we increased the instance importance heterogeneity, we noted a sharp downturn in the performance of GLMM (Fig \ref{fig:simulation}b), a trend also observed in other conventional ML models (Figure~\ref{sfig:simulations}(ii)). In contrast, MIL models maintained their accuracy more effectively.
When varying the number of instances per bag, the number of instance features, and the variance attributed to bag embedding effects, MixMIL consistently outperformed baselines (Figure~\ref{fig:simulation}c, Figure~\ref{sfig:simulations}).
We also compared MixMIL with a version utilizing a mean field posterior. As the latter exhibited slightly diminished performance (Figure~\ref{sfig:simulations}(i) and \ref{sfig:simulations_instance_retrieval}), it was not considered in the real data analyses.
To conclude, in instance retrieval tasks, MixMIL consistently outperformed baseline MILs, accurately retrieving the top simulated instances (Figure~\ref{fig:simulation}d-f).

\subsection{Single-Cell Genomics Dataset}
\paragraph{Task} We here consider the task of predicting genetic variants from transcriptional cell embeddings in the OneK1K dataset~\citep{YazarOneK1K}.
This task carries biological significance: Identifying genetic variants associated with cellular transcriptional states can pinpoint cellular processes implicated in health and disease~\citep{gtex2017,westra2013}.

\paragraph{Dataset}
The OneK1K dataset comprises single-cell RNA sequencing (scRNA-seq) data from approximately 1.3 million peripheral blood mononuclear cells, derived from 982 genotyped donors.
For our analysis, we considered the sub-lymphoid cells (CD4, CD8, NK cells), spanning cells sharing core lymphoid pathways yet exhibiting distinct functionalities.
Regarding genetic labels, we focused on independent variants associated with the average transcriptional state (see next paragraph).
The final dataset for our analysis consisted of 981 individuals (bags), 1.1M cells (instances\footnote{The number of cells per donor ranged from 139 to 2587}), and 28 genetic labels.  

\paragraph{Setup}
Cell state embeddings were obtained from single-cell expression data using single-cell Variational Inference (scVI) with 30 latent factors~\citep{Lopez2018scVI}, a deep generative model for cell state representation learning (see Appendix~\ref{sec:single_cell_embeddings} for full details).
To identify the set of independent genetic labels for our task, we processed 16,718 variants known as cis-eQTLs, using a series of criteria including their association with average transcriptional state and linkage equilibrium considerations. This procedure yielded 28 distinct variants, which we used as labels when comparing alternative MIL implementations.
A detailed methodology on this selection process can be found in Appendix~\ref{sec: variant_filtering_procedure}. For all models, we controlled for sex and age, and additionally accounted for population structure by using the four leading principal components of genetic data. We used MixMIL's fixed effects for this purpose and regressed these covariates from the data for the baseline MILs.
Out-of-sample prediction performance for MIL models was computed utilizing a 5-fold stacked cross-validation procedure.
Briefly, we concatenated the out-of-sample predictions on each test fold, forming a single prediction vector for all samples.
We then correlated this prediction vector with the observed data using Spearman correlation.
%
\begin{table}[htbp]
  \captionsetup{font=small}
  \caption{Running times and number of parameters for MixMIL and baseline MILs on the genetics dataset. Specifically, we report batch training times (ms) and prediction times (ms) benchmarked on a V100 GPU with 32GB memory, alongside counts of likelihood and variational parameters.}
  \label{table:genetics}
  \resizebox{\linewidth}{!}{
    \begin{tabular}{lccccc}
    \toprule
     \textbf{Method} & \textbf{batch time} & \textbf{predict time} & \textbf{lik pars} & \textbf{var pars}\\
       & (ms) & (ms) & (\#) & (\#) \\
    \midrule
    Bayes-MIL  & $5.65 \pm 0.02$ & $16.55 \pm 0.30$ & 2883 & 1890\\ 
    ABMIL   & $1.72 \pm 0.02$ & $0.21 \pm 0.01$ & 1922 & -\\
    Gated ABMIL   & $2.02 \pm 0.02$ & $0.25 \pm 0.01$ & 2852 & -\\
    DSMIL  & $1.99 \pm 0.06$ & $0.39 \pm 0.02$ & 992 & -\\
    \textbf{MixMIL}  & $\textbf{0.14} \pm \textbf{0.01}$ & $\textbf{0.04} \pm \textbf{0.01}$ & 67 & 1890\\
    \bottomrule
    \end{tabular}
  }
\end{table}

\paragraph{Results}
MixMIL demonstrated a consistent enhancement in performance compared to MIL baselines across the majority of the 28 genetic labels (Figure~\ref{fig:genetics}). Specifically, MixMIL outperformed ABMIL for 27 out of 28 labels (P$<2\cdot{10}^{-7}$, from a binomial test; Figure~\ref{fig:genetics}), DSMIL for 25 out 28 labels (P$<2\cdot{10}^{-5}$), Bayes-MIL for 24 out of 28 labels (P$<9\cdot{10}^{-5}$), and Gated ABMIL for 25 out of 28 labels (P$<2\cdot{10}^{-5}$).
Notably, in addition to improved prediction performance, MixMIL showed a marked reduction in running time, being over 12$\times$ faster than ABMIL and over 40$\times$ faster than Bayes-MIL (Table~\ref{table:genetics}).
Relatedly, MixMIL also has a lower complexity than other MIL models, utilizing less than 7\% of the parameters of DSMIL and under 2.4\% of the parameters in Bayes-MIL (Table~\ref{table:genetics}).
Finally, we leveraged MixMIL's instance retrieval to delve deeper into the transcriptional cell states that are most predictive of specific genetic labels.
Notably, some of these states corresponded with known cell types, while others revealed novel biological insights (Figure~\ref{sfig:genetics}).

\subsection{Microscopy Dataset}

\paragraph{Task}
We considered the task to predict a compound's MoA\footnote{
A compound's MoA refers to the specific biological process affected by the compound
} from morphological cell embeddings using the microscopy-based drug screening dataset BBBC021~\citep{Caie2010bbbc021}. Accurate MoA classification is critical in drug discovery as it expedites the understanding of drug effects. As not all cells in culture respond uniformly to the same perturbation, using MIL to model response heterogeneity can improve MoA prediction accuracy and offer insight into phenotypic responses.

\paragraph{Dataset}
The BBBC021 dataset contains microscopy images of MCF7 breast cancer cell lines treated with 113 compounds for 24 hours \citep{Caie2010bbbc021}. Following \cite{ljosa2013comparison}, we focus on 39 compounds with a visible impact on cell morphology, which was associated with 12 distinct MoA labels. The experiments were run on plates with 96 wells, each containing multiple cells. All cells in a well were perturbed by the same compound, and the same compound was used to perturb multiple wells on a single plate and multiple replicate plates. Within this setup, we have 2,526 wells (bags), 133,628 cells (total number of instances), and 12 MoAs (labels). 

\paragraph{Setup}
We extracted morphological cell embeddings from single-cell images using the pre-trained ResNet50 model proposed by \cite{Perakis2021simclr}, which was trained on the same dataset using the self-supervised SimCLR framework~\citep{Chen2020SimCLR}. For all models, we used the leading 256 principal components of the SimCLR embeddings as instance representations---increasing the number of principal components did not significantly improve performance. We, furthermore, accounted for plate batch effects using MixMIL's covariate effects. To accommodate the multiclass classification task, we employed a categorical likelihood for MixMIL as described in Section~\ref{sec:likelihood_choices}. We then compared its performance with the baseline models in Section~\ref{method_considered} using the F1 score metric and balanced accuracy. To evaluate model generalization, we held out one plate per compound for testing and optimized the model on the wells of the remaining plates.
To compute standard errors, we ran three repeat experiments, each time holding out a different plate.
%
\begin{table}[h]
  \captionsetup{font=small}
  \caption{F1 score and balanced accuracy comparison for MixMIL and baseline MILs on the MoA classification task. We report averages and standard errors across three repeat experiments, holding out a different plate per treatment for testing.}
  \label{table: moa_prediction_results}
  \resizebox{\linewidth}{!}{%
    \begin{tabular}{lccc}
      \toprule
      \textbf{Method} & \textbf{Bal. Accuracy} & \textbf{F1 Macro} & \textbf{F1 Micro} \\
      \midrule
      Bayes-MIL & $0.63 \pm 0.02$ & $0.63 \pm 0.02$ & $0.70 \pm 0.01$ \\
      ABMIL & $0.72 \pm 0.02$ & $0.73 \pm 0.01$ & $0.76 \pm 0.01$ \\
      Gated ABMIL & $0.67 \pm 0.03$ & $0.65 \pm 0.03$ & $0.70 \pm 0.03$ \\
      Additive ABMIL & $0.41 \pm 0.00$ & $0.34 \pm 0.00$ & $0.47 \pm 0.02$ \\
      DSMIL & $0.89 \pm 0.02$ & $0.89 \pm 0.02$ & $0.90\pm 0.01$ \\
      \textbf{MixMIL} & $\textbf{0.94} \pm \textbf{0.02}$ & $\textbf{0.94} \pm \textbf{0.01}$ & $\textbf{0.95} \pm \textbf{0.01}$ \\
      \bottomrule
    \end{tabular}
  }
\end{table}

\begin{figure}[h]
    \includegraphics[width=\linewidth]{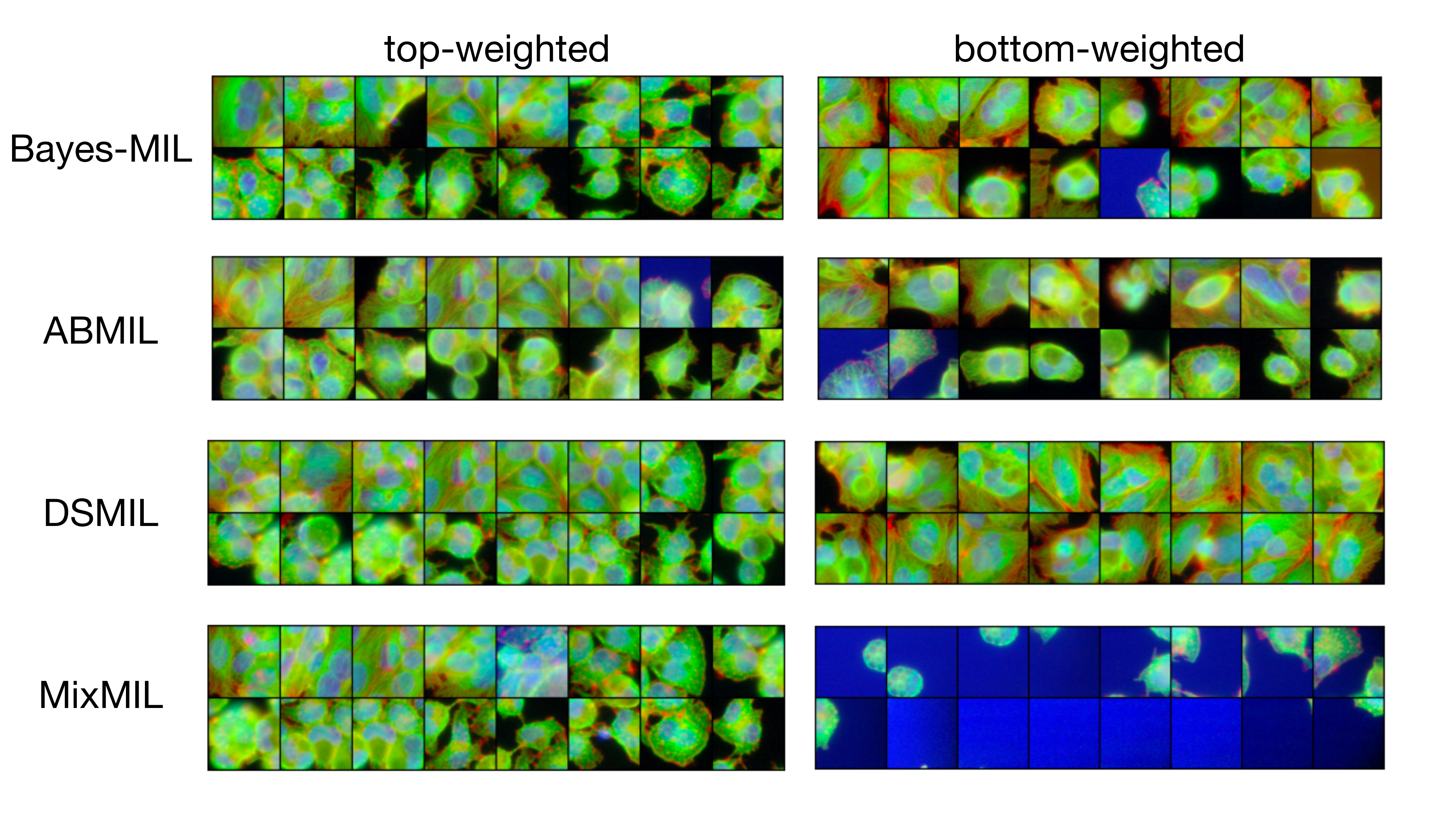}
    \vskip -4mm
    \captionsetup{font=small}
    \captionsetup{skip=20pt}
    \caption{Top and bottom 16 weighted cells for the Latrunculin B drug for different MIL methods.}
    \label{fig:instance_retrieval_microscopy}
\end{figure}

\paragraph{Results}
MixMIL surpassed other MIL models in out-of-sample predictions (Table~\ref{table: moa_prediction_results}). Furthermore, a visual assessment of instances based on MIL methods' attention weights revealed that MixMIL consistently down-weighted experimental artifacts (Figure~\ref{fig:instance_retrieval_microscopy} and Figure~\ref{fig:instance_retrieval_microscopy_sup}).
%

\subsection{Histopathology Dataset}

\paragraph{Task}
We used MixMIL to classify histology slides as cancerous or healthy. Each slide comprises numerous patches, each represented by an embedding.
The application of MIL to this problem enables the assessment of individual patch contributions to the overall slide diagnosis~\citep{Ilse2018attentionmil, ds_mil, bayes_mil}.

\paragraph{Setup and Results}
Our experiments utilized the widely-referenced Camelyon16~\citep{camelyon} histopathology dataset. In particular, we employed the version provided by~\cite{ds_mil}, which offers SimCLR embeddings of 4,886,434 patches at 20x magnification spanning across 399 slides (160 cancerous and 239 healthy), and a predefined train-test split. We earmarked 10\% of the training bags for hyperparameter optimization (see Appendix~\ref{sec:histopath_dataset}). No covariates were available for this dataset. All MIL models exhibited remarkable accuracy, with MixMIL topping the performance chart (Table~\ref{table:histopath}). 
%
\begin{table}[h]
\centering
\captionsetup{font=small}
\caption{AUC classification accuracy results on the test set of the official train-test split of Camelyon16~\citep{ds_mil}, evaluated over five different training seeds.}
\label{table:histopath}
\resizebox{.55\linewidth}{!}{
\begin{tabular}{lc}
\toprule
\textbf{Method} & \textbf{AUC} \\
\midrule
Bayes-MIL & $0.865 \pm 0.031$ \\
ABMIL & $0.958 \pm 0.015$ \\
Gated ABMIL & $0.965 \pm 0.006$ \\
DSMIL & $0.915 \pm 0.013$ \\
\textbf{MixMIL} & $\mathbf{0.977} \pm \mathbf{0.001}$ \\
\bottomrule
\end{tabular}
}
\end{table}

\section{DISCUSSION}
In this work, we presented Mixed Models with Multiple Instance Learning (MixMIL), a framework merging GLMMs and attention-based MIL. Conceived with genomics analyses in mind, MixMIL achieves robustness and efficiency by utilizing pre-trained embeddings and a shallow function for attention modeling.
Our simulations demonstrate the versatility of MixMIL, even in scenarios where simple ML baselines surpassed traditional MIL approaches.
This adaptability was further validated in real-world data applications spanning a wide range of tasks---from applications like genomics, characterized by a lower signal-to-noise ratio, to benchmark MIL datasets in histopathology.
As a limitation,  we note that MixMIL's inherent simplicity could lead to suboptimal performance when working with large datasets on complex tasks.
We hope that by sharing the MixMIL framework, the scientific community finds a valuable tool to analyze multi-instance datasets.
\vspace{-\parskip}

\subsubsection*{Acknowledgments}
We thank the reviewers for their constructive comments. We are grateful to Jose Alquicira-Hernandez for providing access to the processed OneK1K dataset. Special thanks go to Fabiola Curion for her valuable insights and helpful discussions.\newline
JPE received support from the European Laboratory for Learning and Intelligent Systems (ELLIS) through their PhD Program. AP was supported by the Helmholtz Association, as part of the joint research school Munich School for Data Science (MUDS).\newline
Co-funded by the European Union (ERC, DeepCell - 101054957). Views and opinions expressed are however those of the author(s) only and do not necessarily reflect those of the European Union or the European Research Council. Neither the European Union nor the granting authority can be held responsible for them. FJT consults for Immunai Inc., Singularity Bio B.V., CytoReason Ltd, Cellarity, and has ownership interest in Dermagnostix GmbH and Cellarity.\newline
FPC was funded by the Free State of Bavaria's
Hightech Agenda through the Institute of AI for Health (AIH).
\vspace{-\parskip}
\subsubsection*{Contributions}
JPE and FPC developed the model. JPE performed the simulation, single-cell genomics, and histopathology experiments. AP performed the microscopy experiment. JMT provided valuable technical knowledge and contributed to the interpretation of results. FPC conceived the project with support from FJT. FPC and FJT supervised the study. All authors wrote and contributed to the manuscript. The authors read and approved the final manuscript.

\bibliography{bibliography}
\bibliographystyle{plainnat}

\clearpage
\section*{Checklist}
 \begin{enumerate}
 
 \item For all models and algorithms presented, check if you include:
 \begin{enumerate}
   \item A clear description of the mathematical setting, assumptions, algorithm, and/or model. [\underline{Yes}/No/Not Applicable]
   \item An analysis of the properties and complexity (time, space, sample size) of any algorithm. [\underline{Yes}/No/Not Applicable]
   \item (Optional) Anonymized source code, with specification of all dependencies, including external libraries. [\underline{Yes}/No/Not Applicable]
 \end{enumerate}

 \item For any theoretical claim, check if you include:
 \begin{enumerate}
   \item Statements of the full set of assumptions of all theoretical results. [Yes/No/\underline{Not Applicable}]
   \item Complete proofs of all theoretical results. [Yes/No/\underline{Not Applicable}]
   \item Clear explanations of any assumptions. [Yes/No/\underline{Not Applicable}]     
 \end{enumerate}

 \item For all figures and tables that present empirical results, check if you include:
 \begin{enumerate}
   \item The code, data, and instructions needed to reproduce the main experimental results (either in the supplemental material or as a URL). [Yes/\underline{No}/Not Applicable]
   \item All the training details (e.g., data splits, hyperparameters, how they were chosen). [\underline{Yes}/No/Not Applicable]
     \item A clear definition of the specific measure or statistics and error bars (e.g., with respect to the random seed after running experiments multiple times). [\underline{Yes}/No/Not Applicable]
     \item A description of the computing infrastructure used. (e.g., type of GPUs, internal cluster, or cloud provider). [\underline{Yes}/No/Not Applicable]
 \end{enumerate}

 \item If you are using existing assets (e.g., code, data, models) or curating/releasing new assets, check if you include:
 \begin{enumerate}
   \item Citations of the creator If your work uses existing assets. [\underline{Yes}/No/Not Applicable]
   \item The license information of the assets, if applicable. [Yes/\underline{No}/Not Applicable]
   \item New assets either in the supplemental material or as a URL, if applicable. [Yes/\underline{No}/Not Applicable]
   \item Information about consent from data providers/curators. [Yes/No/\underline{Not Applicable}]
   \item Discussion of sensible content if applicable, e.g., personally identifiable information or offensive content. [Yes/No/\underline{Not Applicable}]
 \end{enumerate}

 \item If you used crowdsourcing or conducted research with human subjects, check if you include:
 \begin{enumerate}
   \item The full text of instructions given to participants and screenshots. [Yes/No/\underline{Not Applicable}]
   \item Descriptions of potential participant risks, with links to Institutional Review Board (IRB) approvals if applicable. [Yes/No/\underline{Not Applicable}]
   \item The estimated hourly wage paid to participants and the total amount spent on participant compensation. [Yes/No/\underline{Not Applicable}]
 \end{enumerate}

\end{enumerate}

\appendix
\runningtitle{MixMIL Appendix}
\onecolumn

\setcounter{figure}{0}
\setcounter{equation}{0}
\setcounter{table}{0}
\renewcommand\thefigure{\thesection.\arabic{figure}} 
\renewcommand\thetable{\thesection.\arabic{table}} 
\renewcommand\theequation{\thesection.\arabic{equation}} 
\renewcommand{\thesubfigure}{\roman{subfigure}}
\counterwithin*{table}{section}
\counterwithin*{equation}{section}
\counterwithin*{figure}{section}

\aistatstitle{Mixed Models with Multiple Instance Learning: \\
Appendix}

\section{MixMIL TRAINING DETAILS}
\label{sec:reparameterization}

To enhance numerical stability during training, we reparameterize the embedding-based prediction of our model.
Specifically, introducing a bag-level index $i$ to Eq (3-4) in the main text, we denote the embedding-based bag prediction of bag $i$, composed of $I_i$ instances, as:
%
\begin{equation}
(\B{u})_i = \B{z}_{\B{\gamma}}(\B{X}_i)^T\B{\beta}
\label{eq:bag-pred}
\end{equation}
where $\B{X}_i\in{\mathbb{R}^{I_i\times{Q}}}$ collectively denotes the instance embeddings in bag $i$.
During training, we reparameterize the embedding-based prediction $\B{u}$ to have sample mean 0 and sample variance $b^2 = \frac{1}{Q}\sum_{j=1}^Q\beta_j^2$. Importantly, this reparameterization does not alter the model structure but improves its trainability and interpretability\footnote{After this reparameterization, $\sigma_{\B{\beta}}^2$ can be directly interpreted as the variance explained by pooled bag embeddings.}.
Specifically, we implement the reparameterization by first introducing
%
\begin{equation}
(\B{\Tilde{u}})_i = \B{z}_{\B{\gamma}}(\B{X}_i)^T\underbrace{
\B{\beta}\;/\; b
}_{\B{\eta}},
\label{eq:reparam}
\end{equation}
where $\B{\eta}$ has unit mean square. We then rescale $\B{\Tilde{u}}$ to:
%
\begin{equation}
\B{u} = b \times \frac{\B{\Tilde{u}} - \text{mean}(\B{\Tilde{u}})}{\text{std}(\B{\Tilde{u}})}.
\end{equation}
This standardization operation can be implemented within the Pytorch framework using a \texttt{batchnorm} layer. By using this reparameterization, we introduce stochasticity and stability during training, while keeping track of training set statistics for inference.

\section{BASELINE MODELS}\label{ssec:baseline_models}
\subsection{MIL Baselines}
\begin{itemize}
    \item \textbf{ABMIL.} \cite{Ilse2018attentionmil} use neural networks and attention to learn instance-specific weights and perform bag-level aggregation to optimize a downstream prediction task. The authors additionally propose a gated version of the model where they combine \textit{tanh} and \textit{sigmoid} activation functions to the attention weights to better learn non-linearities.

    \item \textbf{Additive MIL.} \cite{additive_mil} overcome the lack of weight interpretability by pooling instance-level predictions rather than pooling instance features. In greater detail, attention weights are first estimated similarly to ABMIL and used to weigh single instances in a bag. Successively, a predictor is applied directly to the weighted instances to derive instance-specific logits, which are summed to yield a bag-level prediction.  

    \item \textbf{DSMIL.} \cite{ds_mil}  first model an instance classifier which provides class-specific activation scores for each element in a bag. Max-pooling on the classification scores defines a critical instance per class. To obtain bag-level embeddings, attention weights are learned based on the distance of single elements from the critical instance per class and used to aggregate features. 

    \newpage
    \item \textbf{Bayes-MIL.} \cite{bayes_mil} derive uncertainty over the attention weights of a standard MIL model by learning patch-specific posterior distributions with variational inference. In their application to histopathology screens, the authors introduce a slide regularizer to concentrate attention on either the positive or negative side of patches for precise localization with high confidence. Additionally, the authors encode spatial information between patches in Multiple Instance Learning (MIL) for Whole Slide Image (WSI) recognition and localization, proposing the use of Conditional Random Fields (CRF). Because spatial information is not available for the simulation, genetics and microscopy datasets, we use the Bayes-MIL version without slide regularizer and CRF.
\end{itemize}

\subsection{Traditional ML Models}\label{sec:traditional_ml}
\begin{figure}[H]
    \centering
    \subfloat[Comparison with MIL models]{
    \includegraphics[width=0.91\textwidth]{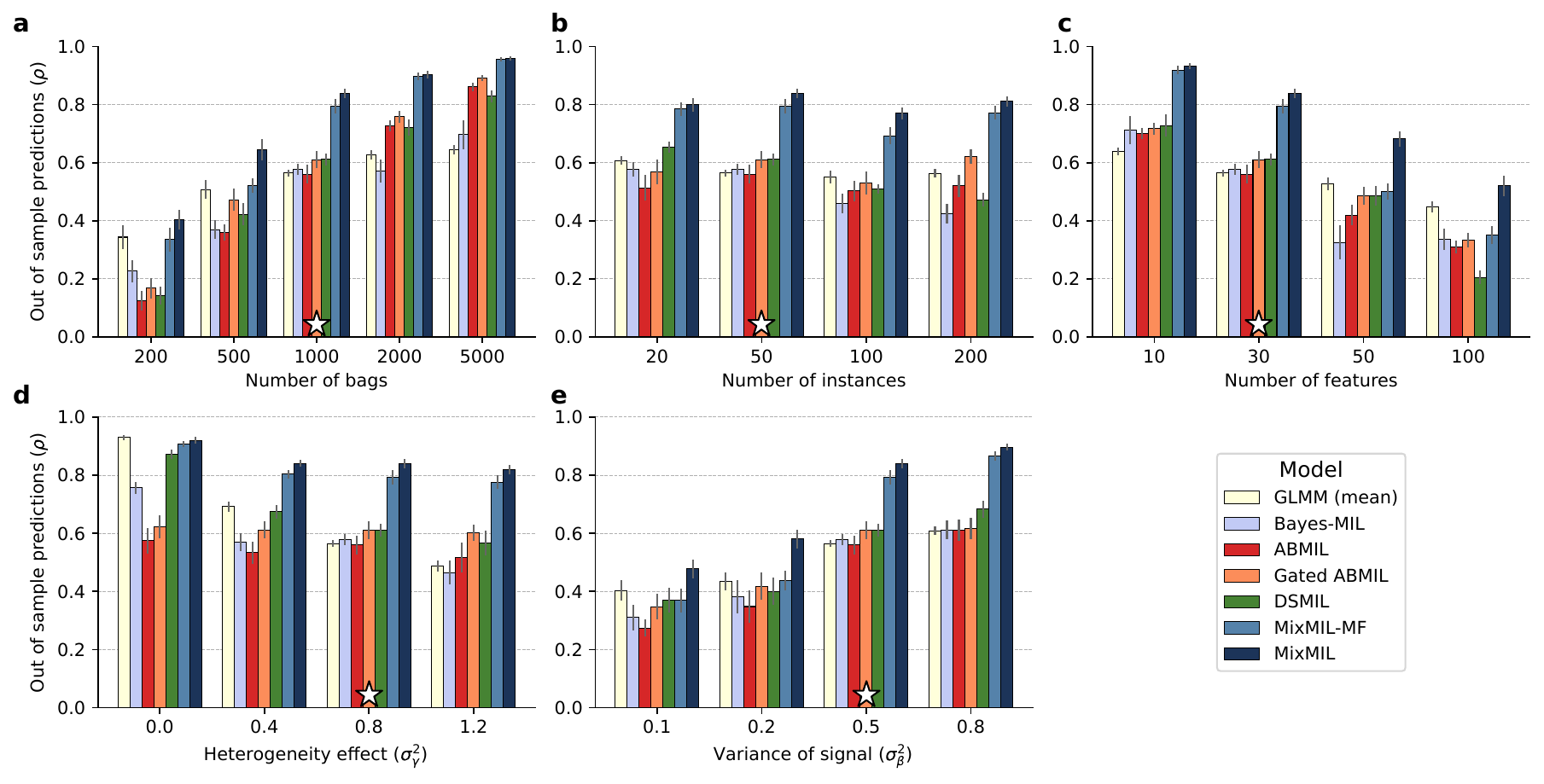}
    }
    \hfill
    \vspace{0.2cm}
    \subfloat[Comparison with vs GLMM, Random Forest, and XGBoost]{
    \includegraphics[width=0.91\textwidth]{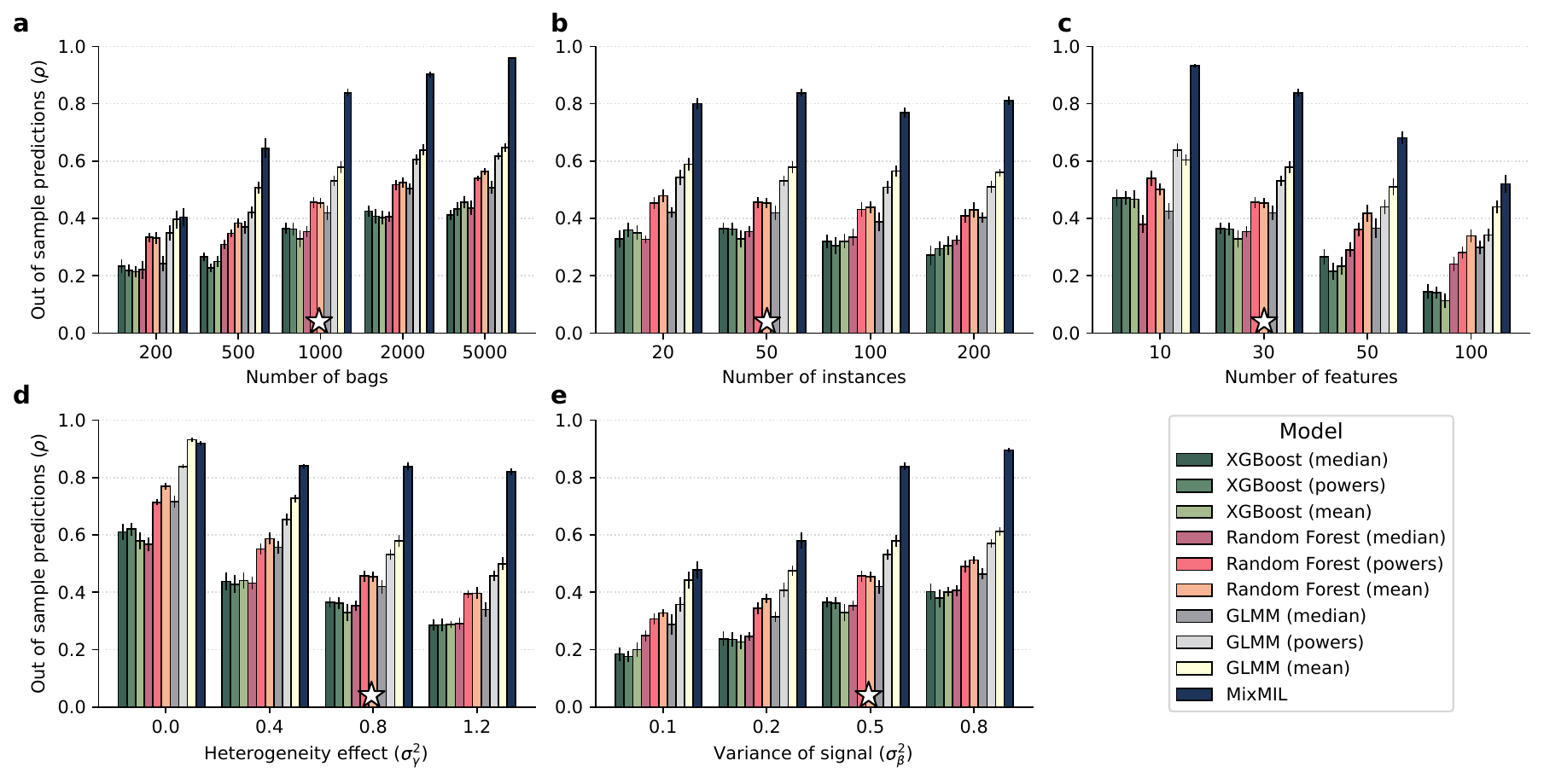}
    }
    \caption{
    Out-of-sample prediction performance (Spearman correlation, $\rho$) in different simulation settings, varying one parameter at a time while keeping the others constant. Specifically, we varied the number of bags (a), the number of instances (b), the number of features (c), the heterogeneity effect (d) and the variance of signal (e).}
    \label{sfig:simulations}
\end{figure}

In simulation studies, we also benchmarked MixMIL against a GLMM with a corresponding likelihood and two widely used nonlinear models: Random Forest and XGBoost. All models use predefined bag-level features as inputs, based on three different strategies: mean pooling, median pooling, and a combination of mean, squared mean, and cubed mean of instance features (Figure~\ref{sfig:simulations}~(ii)).

\section{EXPERIMENTAL SETUP}\label{sec:experimental_setup}
\subsection{Simulations and Genomics Dataset}
\label{single_cell_genomics_dataset}

\paragraph{MIL Model Hyperparameter Search}

Hyperparameters for all MIL models across the simulations and genomics application were optimized based on the simulation default scenario, which emulates the genomics dataset.
Specifically, we used our simulation procedure with default parameters to generate 10 datasets, including training and validation sets.
For each simulated dataset, we trained MIL models with different losses, learning rates and regularization parameters (Table~\ref{tab:sim-hparamsweep}) on the training set and evaluated their performance on the validation set.
The hyperparameters for each model were then chosen based on the average Spearman correlation between predicted and true values, and are listed in Table~\ref{tab:sim-best-params}.
Across the losses that we considered, which included common regression losses (\texttt{MSELoss}, \texttt{HuberLoss}, \texttt{SmoothL1Loss}, \texttt{L1Loss}) as well as the negative log-likelihood of the Binomial distribution used in MixMIL, the \texttt{HuberLoss} consistently yielded the best results for this use case.

%
\begin{table}[H]
\centering
\caption{Hyperparameter sweep for Simulations and Genomics dataset}
\label{tab:sim-hparamsweep}
\resizebox{\textwidth}{!}{%
\begin{tabular}{lllll}
\toprule
Model       & Learning Rate               & Weight Decay        & Dim. Encoder & Regularization  \\
\midrule
Bayes-MIL   & \{5e-3, 5e-4, 1e-4, 1e-5\}  & \{5e-4, 1e-4, 5e-5, 1e-5, 1e-6 \}                    & \{30\}           &   \texttt{log10space}(1e-6, 1e-10)     \\ 
DSMIL       & \{5e-3, 2e-4, 5e-5\}        & \{5e-3, 5e-4, 1e-4\}                    & \{30\}           & -              \\
ABMIL       & \{5e-3, 5e-4, 5e-5\}        & \{5e-3, 5e-4, 1e-4, 1e-5, 0\}                    & \{30\}           & -              \\
Gated ABMIL & \{5e-3, 5e-4, 5e-5\}      & \{5e-3, 5e-4, 1e-4, 1e-5, 0\}                    & \{30\}           & -             \\
\bottomrule
\end{tabular}}
\end{table}

\begin{table}[H]
\centering
\caption{Optimized Hyperparameters for Simulations and Genomics dataset}
\label{tab:sim-best-params}
\begin{tabular}{lllll}
\toprule
Model       & Learning Rate      & Weight Decay       & Dim. Encoder & Regularization \\
\midrule
Bayes-MIL   & 5e-4               & 1e-6               & 30           & 1e-8      \\
DSMIL       & 2e-4               & 5e-3               & 30           & -             \\
ABMIL       & 5e-4               & 1e-4               & 30           & -             \\
Gated ABMIL & 5e-4               & 1e-4               & 30           & -             \\
\bottomrule
\end{tabular}
\end{table}

\paragraph{Traditional ML Models Hyperparameter Search}
To tune the hyperparameters of the baseline models considered in our paper, we performed a randomized search of the hyperparameter space, using a 5-fold cross-validation within the training set and sampling 20 hyperparameter combinations.
Specifically, each combination of hyperparameters was evaluated by the average cross-validated Spearman correlation metric, the combination that provided the best performance was chosen, and the final models were retrained on the entire training set before performing out-of-sample predictions on the test set.
For full information on hyperparameters and their respective search spaces see Table~\ref{tab:random-forest-hparamsweep} and Table~\ref{tab:xgboost-hparamsweep}.
%
\begin{table}[H]
\centering
\caption{Random Forest Hyperparameter Search Space}
\label{tab:random-forest-hparamsweep}
\resizebox{0.55\textwidth}{!}{%
\begin{tabular}{ll}
\toprule
Hyperparameter & Search Space \\
\midrule
Number of Estimators (n\_estimators) & \{50, 100, 150, 200, 500\} \\
Max Features (max\_features) & \{\texttt{sqrt}, \texttt{log2}, \texttt{None}\} \\
Max Depth (max\_depth) & \{10, 20, 30, 50\} \\
Min Samples Split (min\_samples\_split) & \{2, 5, 10\} \\
Min Samples Leaf (min\_samples\_leaf) & \{1, 2, 4\} \\
\bottomrule
\end{tabular}}
\end{table}
%
\begin{table}[H]
\centering
\caption{XGBoost Hyperparameter Search Space}
\label{tab:xgboost-hparamsweep}
\resizebox{0.55\textwidth}{!}{%
\begin{tabular}{ll}
\toprule
Hyperparameter & Search Space \\
\midrule
Max Depth (max\_depth) & \{3, 4, ..., 9\} \\
Learning Rate (learning\_rate) & \{$10^{-3}$, ..., $10^{-2}$, ..., $10^{0}$\} \\
Number of Estimators (n\_estimators) & (100, 1000) \\
Subsample (subsample) & (0.5, 1) \\
Col Sample By Tree (colsample\_bytree) & (0.5, 1) \\
Objective (objective) & \texttt{binary:logistic} \\
Evaluation Metric (eval\_metric) & \texttt{logloss} \\
\bottomrule
\end{tabular}}
\end{table}

\subsection{Microscopy Dataset}
\label{microscopy_dataset}

For the microscopy data, we ran three repeat experiments, each time holding out a different plate as a test set.
For each repeat experiment, we trained MIL models with different hyperparameters on 90\% of the training set (Table~\ref{tab:microscopy-hparamsweep}) and evaluated their predictive performance based on the F1 score on the remaining 10\%.
For each model, we considered hyperparameters that yielded the best performance on this validation set and retrained on the entire training set before computing predictions on the test set.
We noticed that the same hyperparameters were selected for each model across the three repeat experiments (Table~\ref{tab:microscopy-best-params}).

\begin{table}[H]
\centering
\caption{Hyperparameter sweep for Microscopy Dataset}
\label{tab:microscopy-hparamsweep}
\resizebox{0.5\textwidth}{!}{%
\begin{tabular}{lllll}
\toprule
Model       & Learning Rate          & Dim. Encoder  \\
\midrule
Bayes-MIL   & \{1e-3, 5e-4, 1e-4\}  & \{64, 100, 128\}  \\ 
DSMIL       & \{1e-3, 5e-4, 1e-4\}  & \{64, 100, 128\}  \\
ABMIL       & \{1e-3, 5e-4, 1e-4\}  & \{64, 100, 128\}   \\
Additive ABMIL       & \{1e-3, 5e-4, 1e-4\}  & \{64, 100, 128\}   \\
Gated ABMIL & \{1e-3, 5e-4, 1e-4\}  & \{64, 100, 128\}    \\
\bottomrule
\end{tabular}}
\end{table}
%
\begin{table}[H]
\centering
\caption{Optimized Hyperparameters for Microscopy Dataset}
\label{tab:microscopy-best-params}
\begin{tabular}{lllll}
\toprule
Model       & Learning Rate      & Weight Decay       & Dim. Encoder \\
\midrule
Bayes-MIL   & 1e-4               & 1e-5               & 64      \\
DSMIL       & 1e-4               & 1e-5               & 100       \\
ABMIL       & 5e-4               & 1e-5               & 100        \\
Additive ABMIL       & 5e-4               & 1e-5               & 100        \\
Gated ABMIL & 5e-4               & 1e-5               & 100         \\
\bottomrule
\end{tabular}
\end{table}

\subsection{Histopathology Dataset}\label{sec:histopath_dataset}
We employed the version of Camelyon16 provided by~\citep{ds_mil}, which contained a predefined train-test split. We earmarked 10\% of the training bags for hyperparameter optimization and swept MIL model hyperparameters using a grid-search approach (Table~\ref{tab:camelyon-hparamsweep}). For each model, we picked hyperparameters based on the validation loss. The final parameters are shown in Table~\ref{tab:camelyon-best-params}.
\begin{table}[H]
\centering
\caption{Hyperparameter sweep for Camelyon16 Dataset}
\label{tab:camelyon-hparamsweep}
\resizebox{\textwidth}{!}{
\begin{tabular}{lllll}
\toprule
Model       & Learning Rate               & Weight Decay        & Dim. Encoder & Regularization \\
\midrule
Bayes-MIL   & \{5e-4, 1e-4, 5e-5, 1e-5\}  & \{1e-4, 1e-5, 1e-6, 1e-7\}                    & \{30, 60, 120\}           &  \texttt{log10space}(1e-6, 1e-10)      \\ 
DSMIL       & \{1e-4, 2e-4, 5e-5\}        & \{5e-3\}  & \{30, 60, 120\}           & -              \\
ABMIL       & \{1e-4, 5e-4, 5e-5\}        & \{1e-4\}                    & \{30, 60, 120\}           & -              \\
Gated ABMIL & \{1e-4, 5e-4, 5e-5\}        & \{1e-4\}                    & \{30, 60, 120\}           & -             \\
\bottomrule
\end{tabular}}
\end{table}

\begin{table}[H]
\centering
\caption{Optimized Hyperparameters for Camelyon16 Dataset}
\label{tab:camelyon-best-params}
\begin{tabular}{lllll}
\toprule
Model       & Learning Rate      & Weight Decay       & Dim. Encoder & Regularization \\
\midrule
Bayes-MIL   & 5e-4               & 1e-5               & 30           & 1e-8      \\
DSMIL       & 1e-4               & 5e-3               & 60           & -              \\
ABMIL       & 5e-5               & 1e-4               & 60           & -              \\
Gated ABMIL & 5e-4               & 1e-4               & 30           & -             \\
\bottomrule
\end{tabular}
\end{table}

\section{DATA PREPROCESSING}

\subsection{Variant Filtering Procedure}\label{sec: variant_filtering_procedure}
To identify the 28 variants associated with the average transcriptional state, we adopted the following approach:
We started from 16,718 variants associated with proximal gene expression in the primary analysis of this data. These variants are typically known as cis-expression Quantitative Trait Loci (cis-eQTLs). Then, we selected variants that could be predicted from average cell embedding using a GLMM (Spearman $\rho>0.15$).
Next, to mitigate dependencies between genetic labels due to linkage disequilibrium~\citep{slatkin2008linkage}, we undertook a clumping procedure~\citep{purcell2007plink}, which yielded the final set of 28 variants.

\subsection{Single-Cell Embeddings}\label{sec:single_cell_embeddings}
Single-cell RNA-seq data consists of a cell-by-gene matrix of RNA counts per cell. For the OneK1K dataset, we have approximately 1.3 million cells and 32,000 genes~\citep{YazarOneK1K}. However, due to technical bias and experimental dropout, many of these genes are uninformative. Therefore, it is common to subset to the most highly variable genes and perform analyses on embedding space. The single-cell Variational Inference (scVI) model~\citep{Lopez2018scVI} is designed with the properties of this data in mind. We used hyperparameters commonly used for this type of dataset size and report them in Table~\ref{tab:scvi_params}.

\begin{table}[h]
\centering
\caption{Summary of scVI~\citep{Lopez2018scVI} parameters}
\begin{tabular}{@{}ll@{}}
\toprule
Parameter                       & Value                                             \\ \midrule
Layers                          & 2                                                 \\
Batch Size                      & 256                                               \\
Epochs                          & 15                                                \\
Gene Likelihood                 & ZINB                                              \\
Covariate                       &  Sequencing Pool                                  \\
Highly Variable Genes (HVG)     & 5000                                              \\ 
\bottomrule
\end{tabular}
\label{tab:scvi_params}
\end{table}
\newpage
As described in the main, we performed the scVI integration on the sub-lymphoid cells (CD4, CD8, NK cells).

\subsection{Microscopy Image Feature Extraction}
We downloaded cell images from \citet{Caie2010bbbc021}, extracted single cell images using nuclear coordinates made available by \citet{ljosa2013comparison}, and applied plate correction as described in \citet{singh2014}. We then collected the weights of a ResNet50 SimCLR model pre-trained by \citet{Perakis2021simclr} to infer cell embeddings. The extracted embeddings have a dimensionality of 2048, which we reduce using PCA.

\subsection{Accounting for Covariates}
The single-cell genomics and microscopy datasets contain covariates and batches which should not be predictive of the outcome labels but could confound the prediction. We, therefore, accounted for them as fixed effects ($\B{c}^T\B{\alpha}$), leveraging the GLMM structure of MixMIL and regressed them from the data for the baseline MILs. We used the covariates sex, age, and population structure (4 genetic PCs) for the genetics use case. In the microscopy experiment, we accounted for plate-batch effects. We applied constant intercepts for the histopathology and simulation experiments.

\newpage
\section{ADDITIONAL RESULTS}
\subsection{Simulations}\label{sec:simulation_appendix}
\newpage
\begin{figure}[H]
    \centering
    \includegraphics[width=0.91\textwidth]{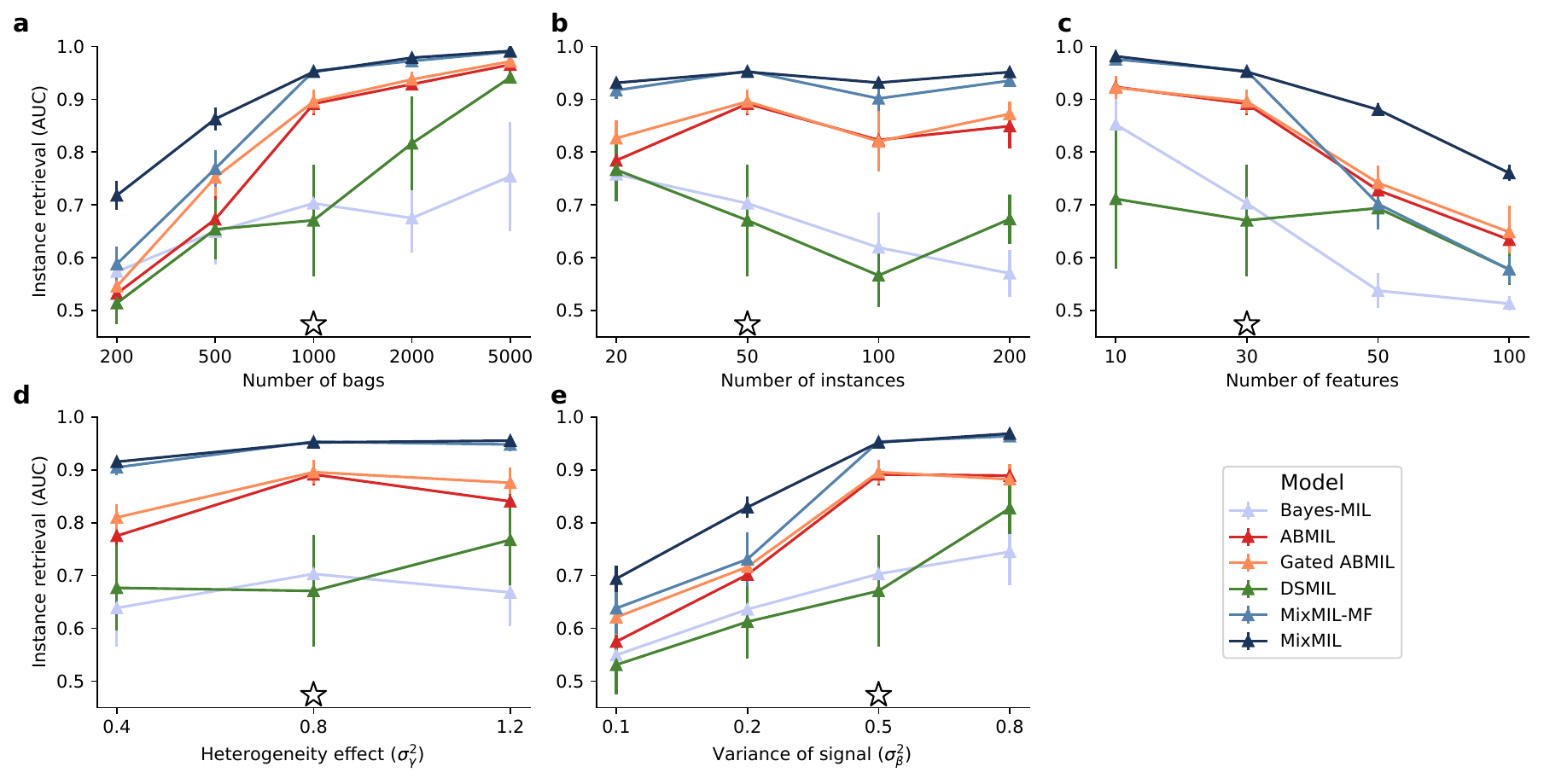}
    \caption{Instance retrieval ROC-AUC of MixMIL for top 10\% of instances in the same simulated scenarios as in Figure~\ref{sfig:simulations}. Error bars denote standard errors across 10 repeat experiments.}
    \label{sfig:simulations_instance_retrieval}
\end{figure}
\begin{table}[H]
\centering
\caption{
Comparison of instance-level models and their mean-embedding-based counterparts with MixMIL in the default scenario (marked by white stars in Figures \ref{fig:simulation}, \ref{sfig:simulations}, and \ref{sfig:simulations_instance_retrieval}). The instance-level models, which either matched or underperformed compared to other models and required longer training times, were not included in further analyses.
}
\label{tab:instance-models}
\begin{tabular}{lcc}
\toprule
Model & Type & Out of sample predictions ($\rho$) \\
\midrule
Random Forest & mean      & $0.45\pm0.02$                        \\
Random Forest & instance-level & $0.45\pm0.02 $                       \\
\midrule
XGBoost & mean   & $0.36\pm0.02$                  \\
XGBoost & instance-level & $0.48\pm0.02 $                       \\
\midrule
GLMM & mean      & $0.58\pm0.02$                   \\
GLMM & instance-level & $0.57\pm0.02$                        \\
\midrule
\textbf{MixMIL}   & MIL & $\textbf{0.84}\pm\textbf{0.01}$                      
\end{tabular}
\end{table}

\subsection{Single-cell Genomics Dataset}

\begin{table}[H]
    \centering
       \caption{Comparison of prediction performance of MixMIL and baseline MILs on the single-cell genomics data. The table reports the mean and standard deviation of Spearman correlation across 28 genes, and the P-values obtained from a paired t-test, assessing the statistical significance of the differences in performance between MixMIL and each of the baseline MILs.}
    \label{tab:genetic_pred_performance}
\begin{tabular}{lcc}
\toprule
Method & Prediction performance (Spearman) & MixMIL improvement (t-test P-value) \\
\midrule
ABMIL & $0.21 \pm 0.18$ & $< 4 \cdot 10^{-7}$ \\
DSMIL & $0.22 \pm 0.17$ & $< 2 \cdot 10^{-7}$ \\
Bayes-MIL & $ 0.23 \pm 0.14$ & $< 2 \cdot 10^{-4}$\\
Gated ABMIL & $0.24 \pm 0.18$ & $< 5 \cdot 10^{-6}$\\
MixMIL & $\textbf{0.29} \pm \textbf{0.17}$ & - \\
\bottomrule
\end{tabular}
\end{table}

\newpage
\begin{figure}[H]
    \centering
    
    \subfloat[\texttt{rs12151621} (Chromosome 2, near GNLY~\citep{GNLY}). Differential gene analysis brought up processes related to defence response (GO:0031349) and regulation of the MAPK cascade (GO:0043408, \cite{MAPK}).]
    {
    \includegraphics[width=0.7\textwidth]{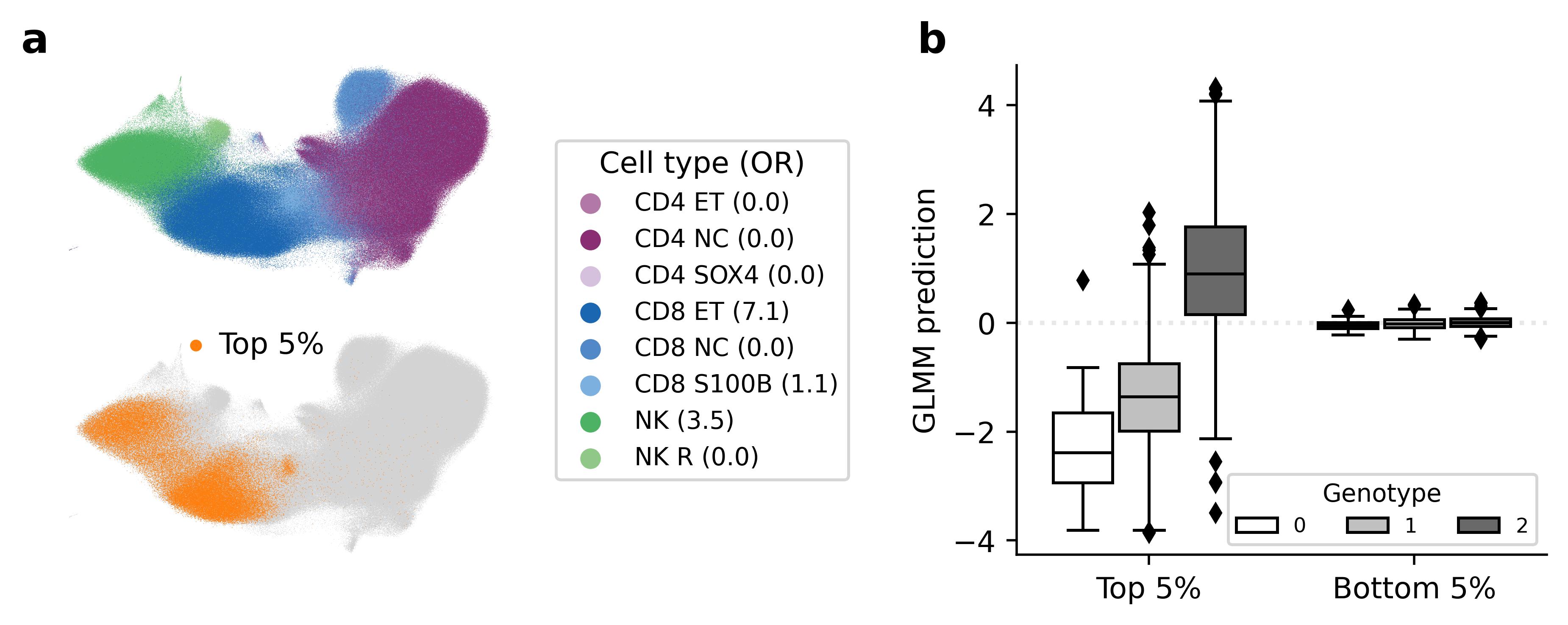}
    }
    
    \hfill
    \vspace{0.6cm}
    
    \subfloat[\texttt{rs9928554} (Chromosome 16, near IL32), associated with Vitamin E measurements (openTargets,~\cite{ochoa2023next}). Differential gene analysis brought up processes related to protein transport to the membrane (GO:0072657) and cell regulatory processes (GO:0050728, GO:0043408, GO:0031349).]{
    \includegraphics[width=0.7\textwidth]{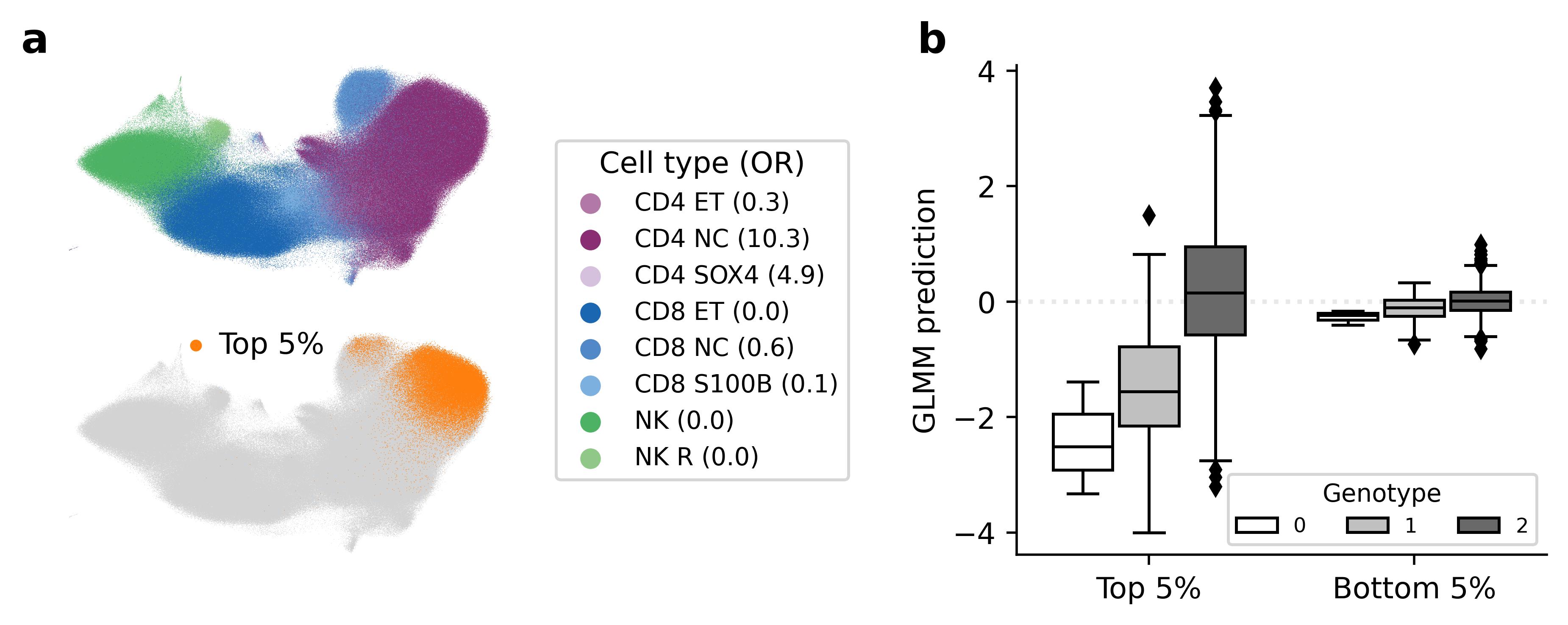}
    }
    
    \hfill
    \vspace{0.6cm}
    
    \subfloat[\texttt{rs7503161} (Chromosome 17, near EIF5A) associated with increased hemoglobin concentration and height (openTargets,~\cite{ochoa2023next}). EIF5A is involved in the positive regulation of the apoptosis (programmed cell death) signaling pathway. Differential gene analysis yielded processes directly (GO:0042981, GO:0043069, GO:0043066) and indirectly (GO:0071345, GO:1902531) related to cell death.]{
    \includegraphics[width=0.7\textwidth]{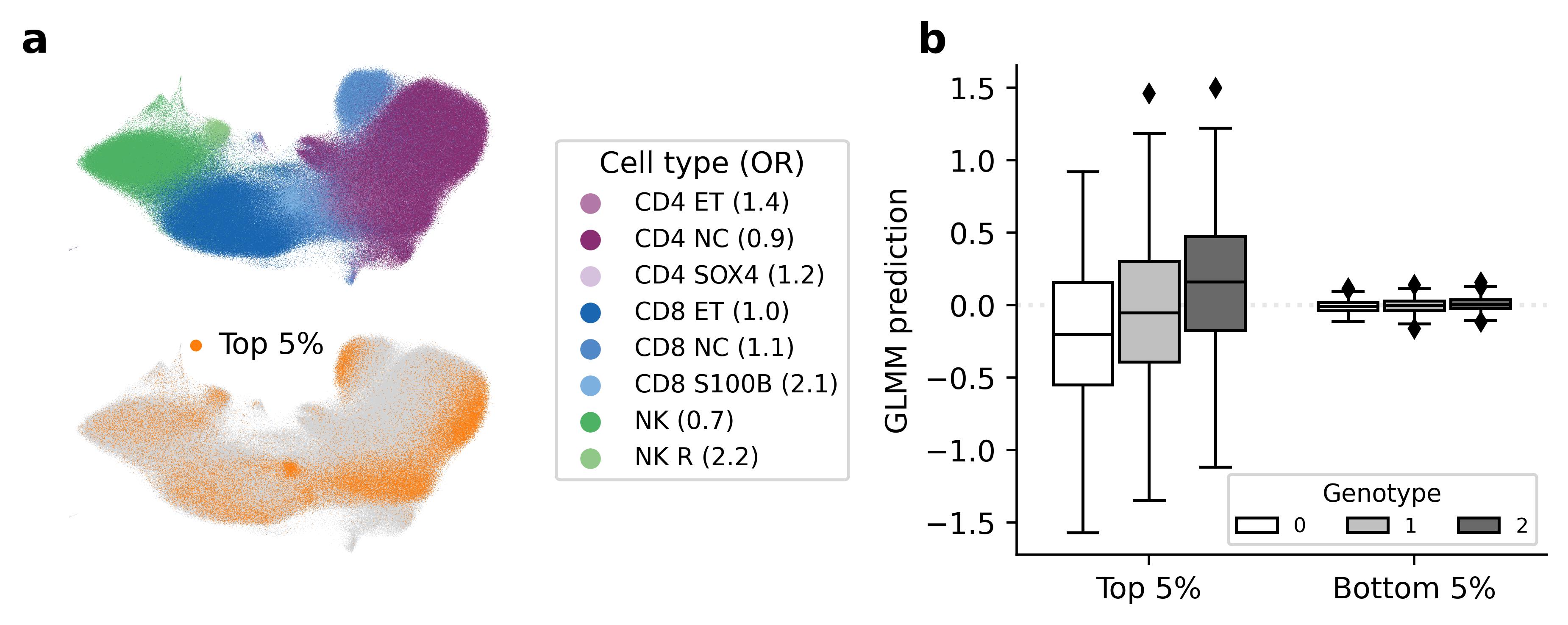}
    }
    \caption{Three examples of genetic variants for which MixMIL improved predictions.
    (a) UMAP of cell transcriptional embeddings showing cell types (top panel) and top 5\% relevant cells according to MixMIL's weights (bottom panel). The odds ratio (OR) quantifies the enrichment of top-weighted cells within each cell type.
    (b) Box plots showing GLMM genetic predictions vs observed values, where the GLMM was fit either using the top 5\% of the instances (left) or the bottom 5\% (right) as ranked by MixMIL.
    Panels (i) and (ii) show variants where MixMIL's importance weights align with known cell types. Conversely, panel (iii) finds a cell state not captured by any individual cell type.}
    \label{sfig:genetics}
\end{figure}

\newpage
\subsection{Microscopy Dataset}
\noindent

\begin{figure}[h]
\begin{centering}
\includegraphics[width=1\textwidth]{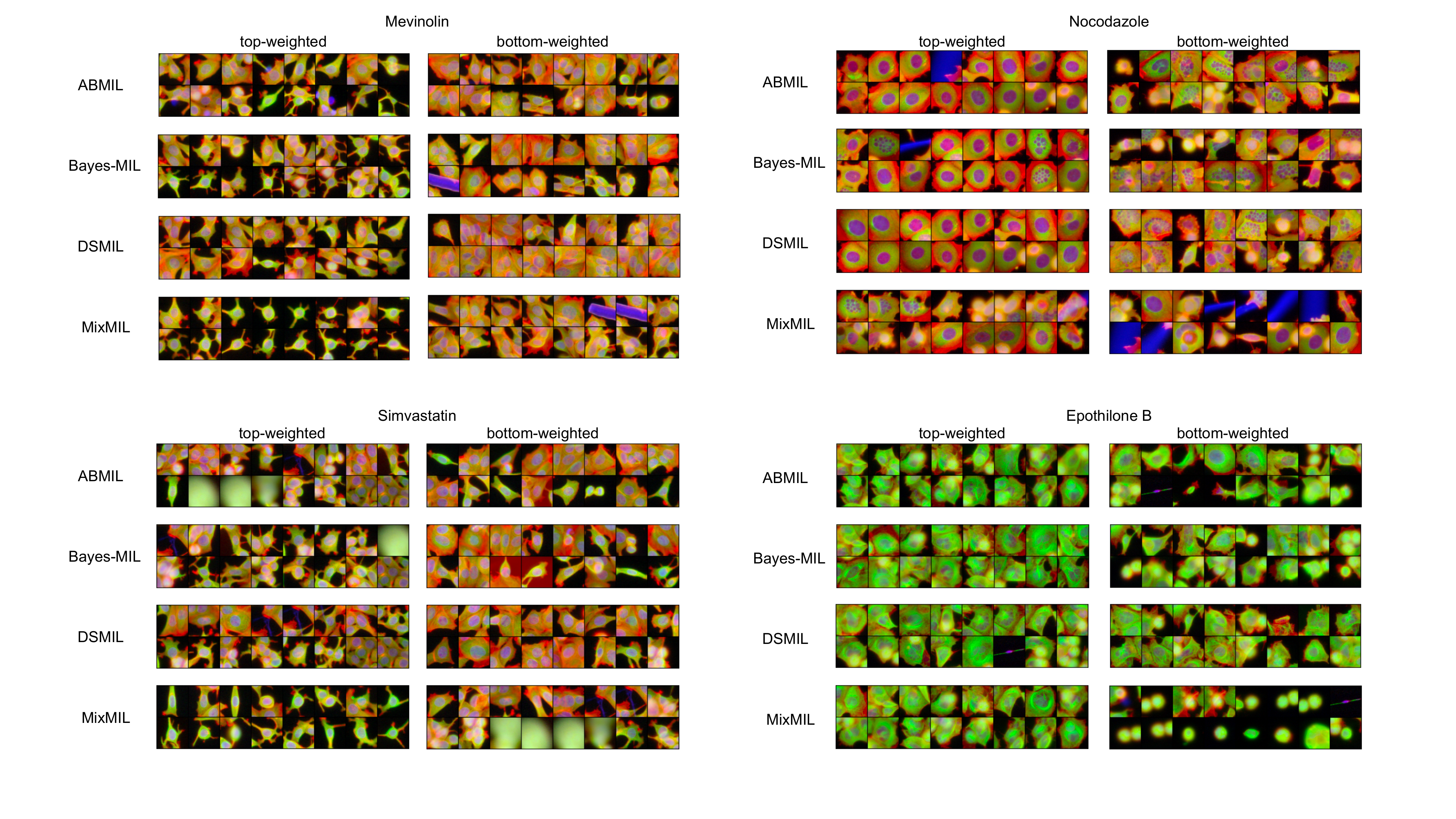}
\vskip -4mm
\captionsetup{font=small}
\captionsetup{skip=20pt}
\caption{Top and bottom 16 weighted cells for the Nocodazole, Simvastatin, Epothilone B, and Mevinolin for different MIL methods.
While MixMIL down-weights images with technical artifacts and yields consistent phenotypes in the top classes, the competing MIL models tend to miss or up-weight noisy images and produce inconsistent phenotypes for the most important perturbed images.
}
\label{fig:instance_retrieval_microscopy_sup}
\end{centering}
\end{figure}

\newpage
\begin{table}[H]
  \captionsetup{font=small}
  \caption{F1 score comparison between MixMIL and competing MIL models on individual MoA labels. We observe that MixMIL improves on the MIL baselines on most considered classes.}
  \label{table:moa_prediction_results}
  \centering
  \resizebox{0.40\textwidth}{!}{
  \begin{tabular}{llc}
    \toprule
    \textbf{MoA Category} & \textbf{Method} & \textbf{F1} \\
    \midrule
    Actin Disruptors & ABMIL & 0.841 ± 0.014 \\
                     & Additive ABMIL & 0.235 ± 0.121 \\
                     & Bayes-MIL & 0.733 ± 0.035 \\
                     & DSMIL & 0.920 ± 0.025 \\
                     & Gated ABMIL & 0.730 ± 0.040 \\
                     & \textbf{MixMIL} & \textbf{0.992 ± 0.004 }\\
    \midrule
    Aurora Kinase Inhibitors & ABMIL & 0.910 ± 0.013 \\
                            & Additive ABMIL & 0.600 ± 0.155 \\
                            & Bayes-MIL & 0.862 ± 0.011 \\
                            & DSMIL & 0.970 ± 0.008 \\
                            & Gated ABMIL & 0.783 ± 0.034 \\
                            & \textbf{MixMIL} & \textbf{1.000 ± 0.000} \\
    \midrule
    Cholesterol-Lowering & ABMIL & 0.550 ± 0.026 \\
                        & Additive ABMIL & 0.562 ± 0.023 \\
                        & Bayes-MIL & 0.293 ± 0.059 \\
                        & DSMIL & 0.885 ± 0.008 \\
                        & Gated ABMIL & 0.415 ± 0.011 \\
                        & \textbf{MixMIL} & \textbf{0.914 ± 0.001 }\\
    \midrule
    DNA Damage & ABMIL & 0.711 ± 0.009 \\
               & Additive ABMIL & 0.316 ± 0.101 \\
               & Bayes-MIL & 0.715 ± 0.027 \\
               & DSMIL & 0.882 ± 0.016 \\
               & Gated ABMIL & 0.558 ± 0.071 \\
               & \textbf{MixMIL} & \textbf{0.922 ± 0.019} \\
    \midrule
    DNA Replication & ABMIL & 0.677 ± 0.013 \\
                    & Additive ABMIL & 0.595 ± 0.015 \\
                    & Bayes-MIL & 0.712 ± 0.012 \\
                    & DSMIL & 0.838 ± 0.030 \\
                    & Gated ABMIL & 0.542 ± 0.039 \\
                    & \textbf{MixMIL} & \textbf{0.928 ± 0.010} \\
    \midrule
    Eg5 Inhibitors & ABMIL & 0.868 ± 0.004 \\
                   & Additive ABMIL & 0.457 ± 0.118 \\
                   & Bayes-MIL & 0.825 ± 0.006 \\
                   & DSMIL & 0.947 ± 0.006 \\
                   & Gated ABMIL & 0.859 ± 0.014 \\
                   & \textbf{MixMIL} & \textbf{0.993 ± 0.004} \\
    \midrule
    Epithelial & ABMIL & 0.759 ± 0.023 \\
               & Additive ABMIL & 0.000 ± 0.000 \\
               & Bayes-MIL & 0.631 ± 0.026 \\
               & DSMIL & 0.865 ± 0.012 \\
               & Gated ABMIL & 0.722 ± 0.026 \\
               & \textbf{MixMIL} & \textbf{0.929 ± 0.008} \\
    \midrule
    Kinase Inhibitors & ABMIL & 0.673 ± 0.040 \\
                      & Additive ABMIL & 0.000 ± 0.000 \\
                      & Bayes-MIL & 0.503 ± 0.048 \\
                      & \textbf{DSMIL} & \textbf{0.956 ± 0.010} \\
                      & Gated ABMIL & 0.658 ± 0.032 \\
                      & MixMIL & 0.927 ± 0.019 \\
    \midrule
    Microtubule Destabilizers & ABMIL & 0.769 ± 0.019 \\
                             & Additive ABMIL & 0.621 ± 0.024 \\
                             & Bayes-MIL & 0.701 ± 0.012 \\
                             & DSMIL & 0.932 ± 0.002 \\
                             & Gated ABMIL & 0.816 ± 0.009 \\
                             & \textbf{MixMIL} & \textbf{0.968 ± 0.006} \\
    \midrule
    Microtubule Stabilizers & ABMIL & 0.932 ± 0.014 \\
                           & Additive ABMIL & 0.283 ± 0.146 \\
                           & Bayes-MIL & 0.858 ± 0.013 \\
                           & DSMIL & 0.964 ± 0.006 \\
                           & Gated ABMIL & 0.822 ± 0.020 \\
                           & \textbf{MixMIL} & \textbf{1.000 ± 0.000} \\
    \midrule
    Protein Degradation & ABMIL & 0.312 ± 0.020 \\
                        & Additive ABMIL & 0.292 ± 0.038 \\
                        & Bayes-MIL & 0.275 ± 0.027 \\
                        & DSMIL & 0.688 ± 0.054 \\
                        & Gated ABMIL & 0.321 ± 0.037 \\
                        & \textbf{MixMIL} & \textbf{0.773 ± 0.026} \\
    \midrule
    Protein Synthesis & ABMIL & 0.757 ± 0.023 \\
                      & Additive ABMIL & 0.076 ± 0.039 \\
                      & Bayes-MIL & 0.447 ± 0.041 \\
                      & DSMIL & 0.879 ± 0.008 \\
                      & Gated ABMIL & 0.625 ± 0.016 \\
                      & \textbf{MixMIL} & \textbf{0.960 ± 0.006} \\
    \bottomrule
  \end{tabular}}
\end{table}

\begin{table}[H]
  \caption{Comparison of MixMIL and MIL baseline models with varying numbers of embedding principal components based on balanced accuracy, F1-macro, and F1-micro, including confidence intervals. Consistent with previous scenarios, MixMIL outperforms other approaches on all evaluated feature dimensions.}
  \label{table:method_comparison_pcs}
  \centering
  \resizebox{0.8\textwidth}{!}{
  \begin{tabular}{cccccc}
      \toprule
    \textbf{Method }& \textbf{Number of Features }& \textbf{Balanced Accuracy} & \textbf{F1-macro} &\textbf{ F1-micro}  \\
    \midrule
    ABMIL & 64 & 0.581 ± 0.036 & 0.577 ± 0.041 & 0.607 ± 0.047 \\
    & 256 & 0.726 ± 0.018 & 0.730 ± 0.016 & 0.764 ± 0.015 \\
    & 512 & 0.714 ± 0.010 & 0.709 ± 0.013 & 0.746 ± 0.009 \\
    \midrule
    Additive ABMIL & 64 & 0.298 ± 0.022 & 0.203 ± 0.020 & 0.340 ± 0.028 \\
    & 256 & 0.410 ± 0.003 & 0.336 ± 0.002 & 0.470 ± 0.019 \\
    & 512 & 0.475 ± 0.045 & 0.379 ± 0.051 & 0.501 ± 0.051 \\
    \midrule
    Bayes-MIL & 64 & 0.727 ± 0.042 & 0.733 ± 0.044 & 0.758 ± 0.031 \\
    & 256 & 0.623 ± 0.018 & 0.628 ± 0.024 & 0.699 ± 0.014 \\
    & 512 & 0.641 ± 0.005 & 0.642 ± 0.010 & 0.693 ± 0.011 \\
    \midrule
    DSMIL & 64 & 0.849 ± 0.028 & 0.850 ± 0.026 & 0.859 ± 0.022 \\
    & 256 & 0.892 ± 0.025 & 0.894 ± 0.023 & 0.902 ± 0.019 \\
    & 512 & 0.890 ± 0.018 & 0.895 ± 0.018 & 0.908 ± 0.014 \\
    \midrule
    Gated ABMIL & 64 & 0.528 ± 0.016 & 0.508 ± 0.016 & 0.582 ± 0.032 \\
    & 256 & 0.669 ± 0.031 & 0.654 ± 0.032 & 0.701 ± 0.031 \\
    & 512 & 0.674 ± 0.029 & 0.676 ± 0.027 & 0.707 ± 0.027 \\
    \midrule
    MixMIL & \textbf{64} & \textbf{0.912} \textbf{±} \textbf{0.019} & \textbf{0.915 ± 0.018} & \textbf{0.924 ± 0.017} \\
    & \textbf{256} & \textbf{0.939 ± 0.017} & \textbf{0.942 ± 0.015} & \textbf{0.950 ± 0.013} \\
    & \textbf{512} & \textbf{0.939 ± 0.019} & \textbf{0.944 ± 0.016} & \textbf{0.951 ± 0.014} \\
    \bottomrule
  \end{tabular}}
\end{table}

\end{document}